\documentclass[11pt]{article}
\usepackage{acl}
\usepackage{times}
\usepackage{latexsym}
\usepackage{amsmath}
\usepackage{amssymb}
\usepackage{graphicx}
\usepackage{booktabs}
\usepackage{array}
\usepackage{makecell}
\usepackage{tabularx}
\usepackage{xcolor}
\usepackage{tikz}
\usepackage{pgfplots}
\usetikzlibrary{positioning,arrows.meta,shapes.geometric,calc}
\pgfplotsset{compat=1.18}

\newcommand{\tasp}{\textsc{TASP}}
\newcommand{\mom}{\textsc{MoM}}

\title{Task-Aware Spectral Pruning: A Mixture-of-Masks Framework for Efficient LLM Inference}

\author{
\textbf{Ibne Farabi Shihab}\textsuperscript{1}
\and
\textbf{Fariya Afrin}\textsuperscript{2}
\and
\textbf{Sanjeda Akter}\textsuperscript{1}
\and
\textbf{Anuj Sharma}\textsuperscript{3}
\\[2pt]
\textsuperscript{1}Department of Computer Science, Iowa State University \\
\textsuperscript{2}Department of Computer Science, Kalinga Institute of Industrial Technology \\
\textsuperscript{3}Department of Civil, Construction \& Environmental Engineering, Iowa State University \\
\texttt{ishihab@iastate.edu}
}

\begin{document}
\maketitle

\begin{abstract}
Static pruning imposes one sparse structure on every prompt, even though reasoning, retrieval, generation, coding, and translation can depend on different parts of a language model. We introduce Task-Aware Spectral Pruning (\tasp), a post-training framework that calibrates module-level spectral descriptors against measured task-specific ablation effects, closes grouped-query-attention and SwiGLU dependencies while constructing sparse masks, and routes each user turn to one compiled mask that remains fixed through prefill and decoding. A module-disjoint pilot determines whether the spectral signal is informative before full calibration. Under the stated retrospective operating rule, the pilot passes on the evaluated Llama-3-8B and Llama-3-70B checkpoints but rejects Qwen2.5-1.5B, showing that applicability is model-dependent rather than universal. At a 43\% active-FLOP reduction, the Llama-3-70B benchmark harness retains $97.7{\pm}0.2\%$ of the dense BF16 score. In the deployment-matched INT8-weight/BF16-compute runtime on one A100 80GB, the compiled sparse path retains $97.3{\pm}0.2\%$ relative to dense BF16 and reduces decode latency from $45.2{\pm}0.4$ to $31.3{\pm}0.4$ ms/token, a $1.44\times$ speedup. Factorized ablations, disjoint-module tests, compiled structured baselines, routing-corruption studies, and an explicit 136-GPU-hour calibration audit delimit both the source and the operating regime of these gains.
\end{abstract}

\section{Introduction}

Large language models support mathematical reasoning, factual retrieval, code generation, translation, and open-ended generation, but their inference cost remains a central obstacle to latency-sensitive and resource-constrained deployment \citep{brown2020language,chowdhery2022palm}. Post-training methods such as SparseGPT, Wanda, LLM-Pruner, SliceGPT, and ShortGPT expose substantial redundancy without repeating pretraining \citep{frantar2023sparsegpt,sun2023wanda,ma2023llmpruner,ashkboos2024slicegpt,men2025shortgpt}. Most of these methods nevertheless commit to one compressed structure, or to a size-indexed family that is agnostic to the input task.

That commitment creates what we call a versatility tax. A mask that preserves arithmetic and code may retain modules different from those needed for translation or fluent generation. Analyses of Transformer circuits suggest functional specialization among attention and feed-forward components \citep{elhage2021mathematical,olsson2022context}, while structured-pruning results show that layers and submodules differ markedly in compressibility \citep{ma2023llmpruner,ashkboos2024slicegpt,lu2024alphapruning}. These observations motivate a more specific question: can a frozen model reveal, before deployment, enough task-dependent structure to build a small family of executable subnetworks rather than a single compromise mask?

Spectral descriptors offer a compact weight-space view of this structure. Heavy-Tailed Self-Regularization relates deviations in empirical weight spectra to learned organization \citep{martin2018implicit,martin2021implicit}, and spectral pruning uses related statistics to allocate sparsity \citep{lu2024alphapruning}. A universal tail threshold, however, need not identify task-sensitive modules across architectures. The useful object is not the threshold itself, but the out-of-sample relationship between a module's spectral profile and the damage caused by removing it for a particular task.

\begin{figure*}[t]
\centering
\begin{tikzpicture}[
  stage/.style={draw, rounded corners=2pt, align=left, text width=.395\textwidth,
    minimum height=2.05cm, inner sep=7pt, font=\small},
  flow/.style={-Latex, line width=.8pt},
  note/.style={font=\scriptsize, align=center}
]
\node[stage, fill=blue!6] (pilot) {
  \emph{1. Profile and test}\\[2pt]
  Head and SwiGLU slices\\
  $\widehat\alpha$, effective rank, norm, tail mass, salience\\
  Structural-ablation labels\\[2pt]
  Gate: pass to full calibration; fail and stop
};
\node[stage, fill=green!7, right=10mm of pilot] (calib) {
  \emph{2. Calibrate and construct}\\[2pt]
  Task-specific ridge predictors\\
  Score per active FLOP\\
  GQA dependency closure\\
  Coupled gate/up/down channels\\[2pt]
  Compile five static mask plans
};
\node[stage, fill=orange!9, below=6mm of calib] (route) {
  \emph{3. Route once per turn}\\[2pt]
  Three-layer prompt router\\
  Temperature-scaled confidence\\[2pt]
  $\max_t p(t\mid x)\ge\tau$:\ sparse mask\\
  otherwise: dense INT8 fallback
};
\node[stage, fill=purple!7, below=6mm of pilot] (exec) {
  \emph{4. Execute consistently}\\[2pt]
  Shared packed weight store\\
  One launch plan for prefill\\
  Same plan for every decode step\\[2pt]
  No cross-mask KV-cache reuse
};
\draw[flow] (pilot) -- node[note, above]{pass} (calib);
\draw[flow] (calib) -- (route);
\draw[flow] (route) -- (exec);
\end{tikzpicture}
\caption{The complete \tasp pipeline. The predeployment gate prevents a universal spectral heuristic from being applied when its out-of-fold signal is weak. A passing model proceeds to task-specific calibration and dependency-closed mask construction. At inference, routing and dispatch occur once per user turn, and the selected compiled plan remains fixed through prefill and decoding so that its KV cache stays valid.}
\label{fig:pipeline}
\end{figure*}
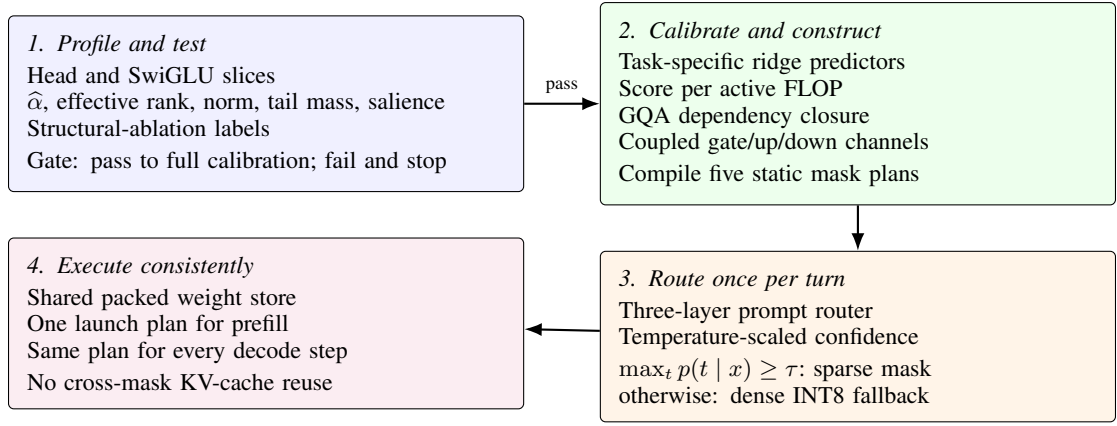

We therefore introduce \tasp, whose stages are summarized in Figure~\ref{fig:pipeline}. A stratified pilot first tests whether low-capacity spectral predictors separate sensitive from insensitive modules on held-out folds. If the model passes, task-specific predictors are fitted to structural-ablation effects, and their scores drive dependency-closed mask construction under an explicit active-FLOP budget. A calibrated router then chooses one precompiled mask for a user turn; low-confidence prompts fall back to the dense path, and the chosen execution plan remains unchanged through prefill and autoregressive decoding.

The contribution is the validated coupling of these stages, not a claim that spectral descriptors, structured pruning, or routing are individually new. First, \tasp turns spectral pruning into an out-of-sample task-sensitivity problem and supplies a numerical rejection rule for unsuitable checkpoints. Second, it connects those predictions to hardware-valid masks by closing grouped-query-attention and SwiGLU dependencies before compilation. Third, it evaluates scoring, mask multiplicity, routing, fallback, compilation, and offline cost separately. This factorization matters empirically: removing spectral features costs 4.1 retention points, replacing five task masks with one global mask costs 4.4 points, and replacing learned masks with layer-matched random masks costs 12.1 points at the same FLOP target.

The evidence also narrows the scope of the method. Spectral--task separability is useful for the evaluated Llama-3-8B and Llama-3-70B checkpoints and near chance for Qwen2.5-1.5B; these experiments do not isolate scale from model family or architecture. Within that scope, the corrected deployment measurements support a quality--latency Pareto claim: ShortGPT is faster in our runtime, whereas \tasp retains substantially more of the dense model. The remainder of the paper develops the method in this order, then evaluates applicability, predictive validity, end-task quality, realized latency, robustness, and amortization.

\section{Related Work}
\label{sec:related}

Post-training pruning ranges from magnitude removal \citep{han2015learning} to second-order reconstruction in SparseGPT \citep{frantar2023sparsegpt} and activation-aware scoring in Wanda \citep{sun2023wanda}. Movement pruning integrates sparsity into fine-tuning \citep{sanh2020movement}, while other work studies stochastic pruning guarantees \citep{anonymous2026certificatepruning}. These approaches often use unstructured masks, which can reduce parameter count without producing smaller regular matrix operations. Structured methods instead remove complete blocks, widths, layers, heads, or coupled channels. LLM-Pruner propagates dependency constraints from gradient-based importance, SliceGPT removes aligned rows and columns, ShortGPT skips complete Transformer blocks, and related methods combine structure with recovery training or elastic execution \citep{ma2023llmpruner,ashkboos2024slicegpt,men2025shortgpt,xia2024shearedllama,gao2024displlm,cai2024flextron,xu2024llamaflex}.

Task-aware pruning derives importance from downstream calibration data rather than a generic language-modeling objective \citep{tian2025taskawarepruning,hong2026dietllm,kodathala2026agent}. \tasp\ shares that goal but does not produce a separate one-task checkpoint. It retains one packed backing store and constructs a small family of static, task-conditioned launch plans. Its predictor is deliberately low-capacity: the experiment asks whether spectral features contain an out-of-sample task-sensitivity signal, not whether a flexible learner can memorize ablation labels.

The closest predecessor in the predicted-importance line is ShadowLLM \citep{bhardwaj2024shadowllm}, which trains a single predictor of \emph{contextual} attention-head and neuron importance from the input prompt and reports executable speedups from the resulting sparsity. \tasp\ and ShadowLLM agree on two design choices---amortizing an importance prediction rather than measuring it at inference, and reporting compiled rather than only analytical savings---and they differ on three. First, ShadowLLM predicts \emph{per-prompt} importance and applies a fresh sparsity pattern to each input, whereas \tasp\ predicts \emph{per-task-family} sensitivity and dispatches one of a small set of precompiled masks per turn, so the executed launch plan is fixed through prefill and decoding and the KV cache stays valid. Second, ShadowLLM's predictor consumes activation features, while \tasp's predictor is fitted to weight-space spectral descriptors and gated by an out-of-fold applicability pilot that can decline to prune a checkpoint. Third, \tasp\ conditions on task rather than token, which is what lets several masks share one packed backing store. We position ShadowLLM as the natural per-prompt counterpart to our per-task design and compare against it directly in Appendix~\ref{app:matched_baselines}; a token-granularity method is not forced onto \tasp's block/head/channel units, and where a fair compiled comparison is infeasible we say so rather than approximate it.

This distinction also separates task specialization from ordinary calibration-based pruning. A task-specialized score can improve one downstream task while sacrificing the breadth expected from a general-purpose model; using one such score globally therefore does not resolve the versatility tax. Conversely, generating several masks is not sufficient if their scores do not identify task-relevant structure. \tasp\ treats these as two independent questions: whether spectral and salience features predict task-specific intervention effects, and whether a router can select among the resulting masks without introducing a larger error than the masks remove. The factorized ablations in Section~\ref{sec:ablation} test both parts under a common FLOP budget.

Random Matrix Theory provides classical reference distributions for high-dimensional spectra \citep{wigner1958distribution,marchenko1967distribution}; trained networks depart from these references in ways associated with learned structure \citep{martin2018implicit,martin2021implicit}. AlphaPruning converts related heavy-tail statistics into layerwise sparsity allocations \citep{lu2024alphapruning}. \tasp\ works at the granularity of attention heads and coupled feed-forward groups, calibrates descriptors against task-specific interventions, and declines to prune when an out-of-fold pilot is uninformative. This differs as well from mixture-of-experts routing, elastic subnetworks, and early exit, which vary token paths, model widths, or inference depth \citep{shazeer2017outrageously,lepikhin2020gshard,cai2024flextron,xu2024llamaflex,xin2020deebert,schuster2021confident}. Further comparisons appear in Appendix~\ref{app:related}.

\section{Task-Aware Spectral Pruning}
\label{sec:method}

Let $\theta$ be a frozen decoder-only Transformer and let $\mathcal{T}$ denote a small set of deployment task families. \tasp\ partitions the model into hardware-valid candidate units, estimates how removing each unit affects each task, and learns a score that can be evaluated for every candidate. A predeployment gate determines whether these scores are informative enough to justify constructing routed masks.
An exploratory taskwise ablation is additionally reported in Appendix~\ref{app:causal}.

\subsection{Candidate Units and Spectral Features}

For a projection matrix or structured slice $W_h\in\mathbb{R}^{d_{\mathrm{in}}\times d_{\mathrm{out}}}$, let $W_h=U\Sigma V^\top$ and define the normalized Gram matrix $C_h=d_{\mathrm{in}}^{-1}W_h^\top W_h$. Its nonzero eigenvalues are $\lambda_i=\sigma_i^2/d_{\mathrm{in}}$. The upper tail above $\lambda_{\min}$ is modeled by a Pareto law,
\begin{equation}
 \Pr(\lambda>x\mid\lambda\ge\lambda_{\min})
 \propto \left(\frac{x}{\lambda_{\min}}\right)^{-\alpha},
\end{equation}
where $\lambda_{\min}$ minimizes the Kolmogorov--Smirnov distance and
\begin{equation}
 \widehat\alpha_h=\left(\frac{1}{n_h}\sum_{i=1}^{n_h}
 \log\frac{\lambda_i}{\lambda_{\min}}\right)^{-1}
\end{equation}
is the maximum-likelihood tail estimate \citep{clauset2009power}. We complement it with spectral norm, normalized layer index, module type, spectral tail-energy ratio
\begin{equation}
 \operatorname{SETOL}(h)=
 \frac{\sum_i\lambda_i\mathbb{1}\{\lambda_i\ge\lambda_{\min}\}}
 {\sum_j\lambda_j},
\end{equation}
and effective rank
\begin{equation}
 r_{\mathrm{eff}}(h)=\exp\!\left(-\sum_i\widetilde\lambda_i
 \log\widetilde\lambda_i\right),\qquad
 \widetilde\lambda_i=\frac{\lambda_i}{\sum_j\lambda_j}.
\end{equation}
These weight-space features are computed from the original BF16 checkpoint by streaming one layer or matrix group at a time; the full BF16 checkpoint is never resident on the timing GPU.

The last feature is a lightweight calibration-time salience statistic. For attention head $(\ell,h)$ with attention matrix $A_{\ell,h}(x)$, let $C(x)$ contain numerical values, operators, logical connectives, quantifiers, and negations, and let $Q(x)$ exclude padding and the fixed system prefix. We define
\begin{equation}
 S_{\ell,h}=\mathbb{E}_{x\sim D_c}\!\left[
 \frac{1}{|Q(x)|}\sum_{q\in Q(x)}\max_{u\in C(x)}A_{\ell,h}(q,u)
 \right].
\end{equation}
For a feed-forward group, the corresponding statistic is the mean absolute activation on the same token set. Salience is only a predictive feature; we do not treat it as a mechanistic attribution.

\subsection{Disjoint Sensitivity Calibration and Applicability Gate}

For task family $t$ and candidate $h$, let $\Delta_h^{(t)}$ be the degradation produced by structurally ablating $h$ while leaving all other candidates active. Reasoning, retrieval, and translation use normalized gold-answer log-likelihood; code and generation use teacher-forced reference-completion negative log-likelihood. Full free-form generation is reserved for final mask evaluation. After per-task normalization, a ridge model predicts
\begin{equation}
 s_h^{(t)}=w_t^\top z_h+b_t,
\end{equation}
where $z_h$ concatenates $\widehat\alpha_h$, $r_{\mathrm{eff}}(h)$,
$\|W_h\|_2$, $\operatorname{SETOL}(h)$, $\ell_h$, $S_h$, and the module-type
indicator.

The candidates are stratified by layer and module type into 70\% training, 15\% validation, and 15\% untouched test modules. Before paying for full labels, \tasp\ uses the first 16 calibration sequences for each validation-module/task pair. Five-fold outer cross-validation is stratified over modules; normalization and the ridge coefficient are selected by inner cross-validation within each outer training split. AUROC is computed only from out-of-fold predictions, with a positive label assigned to the top quintile of measured $\Delta_h^{(t)}$.

The operational gate passes only if the mean point-estimate AUROC is at least 0.62, every diagnostic family is at least 0.58, and the hierarchical-bootstrap 95\% lower confidence bound for the mean is at least 0.60. Reasoning, retrieval, code, and translation enter the gate. Generation labels are collected during the same pass but excluded from the decision because isolated-module AlpacaEval and MT-Bench evaluations are noisier. These thresholds were formalized after the original 70B/Qwen analysis and are therefore a retrospective operating rule, not a universal constant. Once a checkpoint passes, the predictor is refit on training modules, tuned on validation modules, and evaluated once on untouched modules and disjoint prompts.

\subsection{Dependency-Closed Masks and Cache-Consistent Routing}

For each task family, \tasp\ selects retained candidates according to predicted sensitivity per unit of active computation. If $M_h\in\{0,1\}$ indicates retention and $c_h$ is the active-FLOP cost, the idealized objective is
\begin{align}
 M_t^\star&\in\arg\max_{M\in\mathcal{C}}
 \sum_h M_h s_h^{(t)},\\
 &\hspace{8mm}\text{subject to }\sum_h c_hM_h\le B_t.
\end{align}
The implementation greedily adds candidates by score per active FLOP and applies dependency closure before accepting each addition. In grouped-query attention, a KV head remains active whenever any associated query head is retained, and query heads are never reassigned across groups. In a SwiGLU block, the same channel group is retained or removed jointly from $W_{\mathrm{gate}}$, $W_{\mathrm{up}}$, and the matching columns of $W_{\mathrm{down}}$. Embeddings, final normalization, and the language-model head remain dense.

The resulting task masks are packed into contiguous tensors and compiled as static launch plans over one shared INT8 weight store. A three-layer MLP router with hidden widths 512 and 256 predicts a task distribution from lexical features and lightweight task descriptors. Temperature scaling is fitted on a held-out router-validation split. If $\max_t p(t\mid x)<\tau$, the selected deployment policy uses the dense INT8 path; otherwise it dispatches the corresponding sparse plan. The decision is made once per user turn, costs $0.16{\pm}0.02$ ms, and is included in latency. The chosen plan is fixed through prefill and every decode step, and a KV cache is never reused across incompatible masks.

Appendix~\ref{app:procedure} details the complete end-to-end TASP pipeline, including candidate partitioning, labeling budget, compilation, and evaluation protocol.

Appendix~\ref{app:detailed_pipeline} provides detailed pipeline views, illustrating sensitivity prediction, dependency-aware mask construction, and inference-time routing

\section{Experimental Design}
\label{sec:setup}

We evaluate Llama-3-8B, Qwen2.5-1.5B, and Llama-3-70B, using 70B as the primary quality and systems reference. The suite contains reasoning (original MMLU \citep{hendrycks2021mmlu}, GSM8K \citep{cobbe2021gsm8k}, and MGSM \citep{shi2023language}), generation (AlpacaEval ~\citep{alpacaeval} and MT-Bench \citep{zheng2023mtbench}), retrieval (NQ-Open \citep{kwiatkowski2019natural} and TriviaQA \citep{joshi2017triviaqa}), code (HumanEval \citep{chen2021evaluating} and MBPP \citep{austin2021program}), and translation (WMT14 En--De and En--Fr \citep{bojar2014findings}). Because these benchmarks use heterogeneous metrics, we report raw scores and macro-average relative retention,
\begin{equation}
 \operatorname{Ret}(m;r)=\frac{1}{|\mathcal{B}|}
 \sum_{b\in\mathcal{B}}
 \frac{\operatorname{score}_{m,b}}{\operatorname{score}_{r,b}}.
\end{equation}
The dense BF16 checkpoint is the scientific reference $r$ for benchmark claims; the dense INT8 runtime is the systems reference for deployment operating points. This distinction prevents quantization loss from being attributed to pruning.

The comparison covers magnitude pruning, SparseGPT, Wanda, gradient scoring, AlphaPruning, LLM-Pruner, SliceGPT, ShortGPT, LLaMaFlex, and an agent-guided rule; matched input-dependent and task-aware structured baselines (PuDDing, IG-Pruning, Instruction-Following Pruning, and TaBP) are compared under the same runtime in Appendix~\ref{app:matched_baselines}. \tasp-Single uses one global calibrated mask, \tasp-\mom\ uses five masks and the learned router, and \tasp-Oracle supplies the true task family. The principal target retains 0.57 of active model FLOPs, corresponding to a 43\% reduction.

All structured systems are selected on their method-native validation grids before deployment conversion. ShortGPT retains complete blocks, SliceGPT retains its selected reduced width, and LLM-Pruner retains its dependency-aware architecture; we do not alter these structures to favor the \tasp\ compiler. Each selected model is then packed into the same groupwise INT8-weight/BF16-compute runtime and reevaluated after compilation. Unstructured methods remain quality baselines at matched analytical active FLOPs because applying their elementwise masks to the dense kernels does not create an executable smaller model. The exact \tasp\ mask-as-zero control makes this distinction explicit by holding mask quality fixed while removing physical packing.

For stochastic \tasp\ variants, quality is mean $\pm$ standard deviation over five independent calibration, mask, and router runs. For baselines that are deterministic after validation-based artifact selection, uncertainty is one paired-bootstrap standard error over evaluation instances. Latency is mean $\pm$ standard deviation over five independent timing blocks, each containing 200 runs after 50 warmups. Table~\ref{tab:configuration} consolidates the complete configuration; Appendix~\ref{app:statistics} gives the remaining statistical and logging details.

\begin{table*}[t]
\centering
\scriptsize
\begin{tabularx}{\textwidth}{@{}p{.135\textwidth}X p{.135\textwidth}X@{}}
\toprule
Component & Configuration & Component & Configuration \\
\midrule
70B candidates & 80 layers; 64 query-head candidates and 28 coupled 1,024-channel SwiGLU groups per layer; 7,360 total & Module split & Stratified by layer and type: 70/15/15\% train/validation/untouched test, or 5,152/1,104/1,104 modules \\
Pilot & Validation modules; first 16 sequences per module--task pair; stratified five-fold outer module CV & Ablation labels & 32 sequences per train/validation module--task pair; 16 per untouched-test pair; batch size 8 \\
Targets & Normalized gold-answer log-likelihood for reasoning, retrieval, translation; teacher-forced completion NLL for code, generation & Feature preprocessing & Per-task z-normalization fitted only on the modules available to that fitting split \\
Predictor & Ridge regression; coefficient grid $\{10^{-4},10^{-3},10^{-2},10^{-1},1,10,10^2\}$ & Ridge selection & Lowest validation MSE; ties resolved by validation AUROC \\
Mask objective & Greedy predicted sensitivity per active FLOP with dependency closure; 0.57 active-FLOP target & Dependencies & Retain a KV head with any associated query head; jointly retain/remove gate, up, and down channels in groups of 1,024 \\
Router & Three-layer MLP with hidden widths 512 and 256 & Router optimization & AdamW; learning rate $3\!\times\!10^{-4}$; weight decay 0.01; batch 64; at most 20 epochs; patience 3 \\
Fallback & Temperature scaling; $\tau=0.85$, selected for maximum validation retention subject to at least 75\% sparse-path usage; dense fallback & Weight and compute & Symmetric groupwise INT8 weights, group size 128; fused dequantization; BF16 scales, accumulation, activations, attention, and KV cache \\
Compilation & Contiguous retained-head/channel tensors; one static Triton plan per mask; one shared packed weight store & Timing & One A100 80GB SXM; batch 1; 2,048-token prefill; 256-token greedy decode; 50 warmups; 200 measurements \\
Timing boundary & Router, dispatch, prefill, decode, and terminal synchronization included; loading, packing, and one-time compilation excluded & Software & PyTorch 2.4; CUDA 12.1; FlashAttention 2.5; Transformers 4.44; Triton 3.0 \\
\bottomrule
\end{tabularx}
\caption{Complete calibration, construction, routing, and runtime configuration. The 70B model uses INT8 weight-only residency with BF16 compute; it does not use BF16 weight residency, tensor parallelism, CPU offload, or NVMe offload.}
\label{tab:configuration}
\label{tab:protocol}
\end{table*}

\section{Results}
\label{sec:results}

The evaluation follows the pipeline itself. We first ask whether the spectral signal is strong enough to proceed, then test whether it generalizes to unseen modules and prompts, and only then measure end-task retention and realized serving performance.

\subsection{Applicability Is Model-Dependent}
\label{sec:applicability}

Table~\ref{tab:gate} reports the pilot before full calibration. Llama-3-8B and Llama-3-70B satisfy all three conditions, while Qwen2.5-1.5B fails them by a wide margin. Direct 8B deployment at the 43\% FLOP reduction retains $96.4{\pm}0.3\%$ across the full eleven-benchmark suite, between \tasp-Single at $92.9{\pm}0.4\%$ and the oracle at $97.2{\pm}0.2\%$. Thus, the positive evidence is not confined to one 70B checkpoint. The Qwen result nevertheless rules out a claim that decoder-only architecture or parameter count alone makes \tasp\ applicable.

\begin{table}[t]
\centering
\scriptsize
\setlength{\tabcolsep}{3pt}
\begin{tabular}{lcccc}
\toprule
Model & Mean & Minimum & Mean LCB & Gate \\
\midrule
Qwen2.5-1.5B & 0.53 & 0.51 & 0.50 & reject \\
Llama-3-8B & 0.66 & 0.63 & 0.62 & pass \\
Llama-3-70B & 0.78 & 0.74 & 0.75 & pass \\
\bottomrule
\end{tabular}
\caption{Out-of-fold pilot AUROC. ``Minimum'' is the lowest diagnostic-family AUROC, and ``Mean LCB'' is the hierarchical-bootstrap 95\% lower bound for the mean.}
\label{tab:gate}
\end{table}

The final 70B predictor is evaluated on untouched modules and disjoint prompts. AUROC changes from 0.82 to 0.80 for reasoning, 0.76 to 0.72 for retrieval, 0.79 to 0.76 for code, and 0.74 to 0.73 for translation, reducing the mean from 0.78 to 0.75. More importantly for pruning, removing groups ranked least important by \tasp\ causes an average $2.6{\pm}0.3$-point loss on unseen modules, compared with $4.8{\pm}0.3$ for Wanda and $7.4{\pm}0.4$ for layer-matched random selection. The modest train-to-test AUROC change and the intervention result support out-of-sample prediction, but not mechanistic identification. Appendix~\ref{app:diagnostics} provides the full diagnostic view.

\subsection{Quality and Realized Latency}
\label{sec:quality_latency}

The benchmark harness and timed runtime answer related but different questions. In the BF16 scientific harness, \tasp-\mom\ retains $97.7{\pm}0.2\%$ at the 43\% active-FLOP reduction, versus $93.3{\pm}0.3\%$ for one global spectral mask and $94.0{\pm}0.3\%$ for the strongest comparison baseline. The oracle reaches $98.7{\pm}0.2\%$, leaving a one-point routing gap. Appendix~\ref{app:benchmarks} reports benchmark-level scores. Table~\ref{tab:main} reports the main benchmark comparison of TASP against structured pruning baselines at the matched 43\% active-FLOP reduction.

\begin{table*}[t]
\centering
\scriptsize
\resizebox{\textwidth}{!}{%
\begin{tabular}{lcccccccc}
\toprule
Method & MMLU & GSM8K & MGSM & AlpacaEval & NQ-Open & HumanEval & WMT14 avg. & Full retention \\
\midrule
Dense BF16 & 79.2 & 87.5 & 74.3 & 82.1 & 68.4 & 72.6 & 28.3 & 100.0 \\
Wanda & $73.8{\pm}0.3$ & $77.1{\pm}0.5$ & $68.2{\pm}0.4$ & $75.6{\pm}0.4$ & $64.3{\pm}0.3$ & $61.2{\pm}0.5$ & $25.6{\pm}0.2$ & $90.7{\pm}0.3$ \\
SliceGPT & $75.4{\pm}0.3$ & $79.6{\pm}0.5$ & $69.9{\pm}0.4$ & $77.2{\pm}0.3$ & $65.8{\pm}0.3$ & $64.2{\pm}0.5$ & $26.5{\pm}0.2$ & $93.2{\pm}0.3$ \\
LLaMaFlex & $75.8{\pm}0.3$ & $80.2{\pm}0.5$ & $70.4{\pm}0.3$ & $77.9{\pm}0.3$ & $66.1{\pm}0.2$ & $64.8{\pm}0.4$ & $26.9{\pm}0.2$ & $94.0{\pm}0.3$ \\
\tasp-Single & $75.1{\pm}0.3$ & $79.8{\pm}0.4$ & $70.4{\pm}0.3$ & $77.3{\pm}0.3$ & $65.9{\pm}0.2$ & $63.4{\pm}0.4$ & $26.4{\pm}0.2$ & $93.3{\pm}0.3$ \\
\tasp-\mom & $77.9{\pm}0.3$ & $85.2{\pm}0.4$ & $72.8{\pm}0.3$ & $80.4{\pm}0.3$ & $67.2{\pm}0.2$ & $69.8{\pm}0.5$ & $27.6{\pm}0.2$ & $97.7{\pm}0.2$ \\
\tasp-Oracle & $78.4{\pm}0.2$ & $86.1{\pm}0.3$ & $73.5{\pm}0.3$ & $81.3{\pm}0.2$ & $67.8{\pm}0.2$ & $70.9{\pm}0.4$ & $28.1{\pm}0.2$ & $98.7{\pm}0.2$ \\
\bottomrule
\end{tabular}}
\caption{Representative Llama-3-70B benchmark results in the dense-BF16 scientific harness at 43\% active-FLOP reduction. The reasoning column is original four-choice five-shot MMLU \citep{hendrycks2021mmlu}; a separate genuine MMLU-Pro evaluation is reported in Appendix~\ref{app:mmlupro}. Full retention additionally includes MT-Bench, TriviaQA, MBPP, and both WMT14 directions. \tasp\ uncertainty is SD across five independent calibration and mask runs; deterministic-baseline uncertainty is one paired-bootstrap SE over evaluation instances. The complete comparison appears in Appendix~\ref{app:benchmarks}.}
\label{tab:main}
\end{table*}

The advantage of task-conditioned masks is largest on GSM8K and HumanEval, where a global compromise removes components that are valuable for structured generation. The smaller gaps on NQ-Open and translation show that the gain is not uniform, which is why retention is computed from all eleven benchmark-level ratios rather than from the displayed subset. Across the 20--70\% active-FLOP sweep, \tasp-\mom\ remains above the global mask throughout the practically useful 30--60\% region; the oracle gap widens only as pruning becomes aggressive. Appendix~\ref{app:pareto} reports this frontier.

Table~\ref{tab:systems} moves to the deployment-matched INT8-weight/BF16-compute runtime and compiles every structured baseline through the same serving path. Dense INT8 retains $99.6{\pm}0.1\%$ relative to dense BF16. The compiled \tasp-\mom\ sparse path retains $97.3{\pm}0.2\%$ relative to BF16, or $97.7{\pm}0.2\%$ relative to dense INT8, while reducing decode latency from $45.2{\pm}0.4$ to $31.3{\pm}0.4$ ms/token. Its $1.44\times$ speedup is smaller than ShortGPT's $1.63\times$, but its BF16-relative retention is 5.8 points higher. The supported systems result is therefore a quality--latency Pareto improvement, not a claim that \tasp\ is always the fastest structured model.

\begin{table*}[t]
\centering
\scriptsize
\resizebox{\textwidth}{!}{%
\begin{tabular}{lccccccc}
\toprule
Method & Selected density & Active FLOPs & Ret. vs. BF16 & Ret. vs. INT8 & Prefill (ms) & Decode (ms/tok.) & Speedup \\
\midrule
Dense INT8 runtime & $1.00\times$ & $1.00\times$ & $99.6{\pm}0.1$ & $100.0$ & $1{,}814{\pm}18$ & $45.2{\pm}0.4$ & $1.00\times$ \\
\tasp-\mom\ mask-as-zero & $0.57\times$ & $\approx1.00\times$ & $97.3{\pm}0.2$ & $97.7{\pm}0.2$ & $1{,}806{\pm}19$ & $43.8{\pm}0.5$ & $1.03\times$ \\
ShortGPT, compiled & $0.57\times$ & $0.57\times$ & $91.5{\pm}0.4$ & $91.9{\pm}0.4$ & $1{,}118{\pm}15$ & $27.8{\pm}0.3$ & $1.63\times$ \\
LLM-Pruner, compiled & $0.57\times$ & $0.57\times$ & $92.0{\pm}0.3$ & $92.4{\pm}0.3$ & $1{,}371{\pm}17$ & $35.7{\pm}0.4$ & $1.27\times$ \\
SliceGPT, compiled & $0.57\times$ & $0.57\times$ & $92.8{\pm}0.3$ & $93.2{\pm}0.3$ & $1{,}289{\pm}16$ & $33.8{\pm}0.4$ & $1.34\times$ \\
\tasp-Single, compiled & $0.57\times$ & $0.57\times$ & $93.0{\pm}0.3$ & $93.4{\pm}0.3$ & $1{,}342{\pm}16$ & $31.9{\pm}0.4$ & $1.42\times$ \\
\tasp-\mom, compiled sparse path & $0.57\times$ & $0.57\times$ & $97.3{\pm}0.2$ & $97.7{\pm}0.2$ & $1{,}326{\pm}15$ & $31.3{\pm}0.4$ & $1.44\times$ \\
\bottomrule
\end{tabular}
}
\caption{Deployment-matched Llama-3-70B results on one A100 80GB. Active model FLOPs are analytical counts implied by the launched architecture, not hardware-counter measurements. Retention uncertainty for \tasp\ rows is SD across five independently calibrated systems; uncertainty for deterministic baselines is one paired-bootstrap SE. Latency uncertainty is SD across five independent 200-run timing blocks for every row. The mask-as-zero control uses the exact \tasp-\mom\ masks in dense kernels, separating mask quality from compiled execution.}
\label{tab:systems}
\label{tab:latency}
\end{table*}

The all-sparse path is distinct from the selected fallback policy. At $\tau=0.85$, 76${\pm}$3\% of requests use the sparse path and retention rises to $99.3{\pm}0.1\%$ relative to dense INT8. Combining the measured dense and sparse latencies gives an implied 34.6 ms/token and $1.30\times$ speedup. This is a mixture calculation, not a separately timed kernel result. The full operating curve appears in Appendix~\ref{app:robustness}.

\subsection{Spectral Scoring and Mask Diversity Are Complementary}
\label{sec:ablation}

Table~\ref{tab:ablation} follows the causal chain from score to mask to route. The learned-to-oracle gap is only 1.0 point, so perfect routing is not the primary source of the improvement. Removing salience costs 2.2 points and removing spectral features costs 4.1 points. Replacing five masks with one global spectral mask costs 4.4 points, while five random masks lose 12.1 points. Mask multiplicity alone is therefore insufficient: calibrated ranking finds useful subnetworks, and routing selects among task-conditioned alternatives. Because the audit already measures every candidate--task intervention, Appendix~\ref{app:direct_ranking} also compares the spectral predictor against a \emph{direct} measured-ablation ranking at matched label budgets: \tasp\ closes about $77\%$ of the gap to an all-label direct oracle and is close to, but not equal to, exhaustive ranking at a fraction of the label cost.

\begin{table}[t]
\centering
\scriptsize
\setlength{\tabcolsep}{3pt}
\resizebox{\columnwidth}{!}{%
\begin{tabular}{lc}
\toprule
Variant & Retention vs. dense BF16 \\
\midrule
Five masks, oracle router & $98.7{\pm}0.2$ \\
Five masks, learned router & $97.7{\pm}0.2$ \\
Without salience & $95.5{\pm}0.3$ \\
Without spectral features & $93.6{\pm}0.4$ \\
One global spectral mask & $93.3{\pm}0.3$ \\
Five layer-matched random masks & $85.6{\pm}0.5$ \\
\bottomrule
\end{tabular}
}
\caption{Factorized 70B ablation at the same 43\% active-FLOP reduction. Values are mean $\pm$ SD across five calibration and mask runs.}
\label{tab:ablation}
\label{tab:ablation_ladder}
\end{table}

The mask family is neither identical nor uniquely determined. Across task families, reasoning/code overlap is $0.78{\pm}0.02$ and generation/translation overlap is $0.81{\pm}0.02$, whereas reasoning/generation overlap is only $0.24{\pm}0.03$. Within a task, perturbing equal-score ties and the final 5\% of ranked selections produces 10--18 distinct masks within 0.5 retention points, with Jaccard overlap between 0.78 and 0.84. \tasp\ thus identifies a stable core plus replaceable modules rather than a unique circuit. Appendix~\ref{app:masks} reports both forms of overlap.

Routing and calibration degrade smoothly rather than failing at a single threshold. Corrupting top-1 route accuracy from 100\% to 70\% lowers no-fallback retention from $98.7{\pm}0.2$ to $95.2{\pm}0.4$; calibrated dense fallback recovers the latter to $97.5{\pm}0.3$. Using only half of the training module--prompt labels yields held-out AUROC $0.70{\pm}0.02$ and retention $96.7{\pm}0.3$, compared with $0.75{\pm}0.02$ and $97.7{\pm}0.2$ with all labels. These controlled curves, along with task-switching results, appear in Appendix~\ref{app:robustness}. The router's \emph{natural} error rate, its calibration, and its degradation under prompt-, template-, dataset-, and domain-disjoint splits are measured separately in Appendix~\ref{app:router_generalization}, where source-disjoint accuracy falls from $92.4\%$ to $75.6\%$ but the fallback policy keeps retention at $97.5\%$ or above.

\subsection{Calibration Cost and Serving Regime}
\label{sec:cost}

The applicability pilot limits wasted work but does not make \tasp\ cheap. Llama-3-70B contains 7,360 structured candidates, yielding 36,800 module--task ablations across five families and 1,089,280 prompt--module evaluations under the complete split. Spectral profiling, label collection, disjoint testing, predictor fitting, five-mask compilation, and router calibration total 136.0 A100 GPU-hours, compared with 7.4--19.6 hours for the post-training baselines. An unsuitable checkpoint can be rejected after about 15.8 hours, including spectral profiling and the pilot.

For the measured 2,048-token prefill and 256-token decode, the compiled sparse path saves approximately 4.05 seconds per request and amortizes 136 GPU-hours after about 121,000 requests, or 31.0 million generated tokens. The selected $\tau=0.85$ policy saves approximately 3.08 seconds per request and breaks even after about 159,000 requests, or 40.8 million generated tokens. These calculations define the intended regime: sustained serving with stable task families, not one-off inference or frequently changing task definitions. Appendix~\ref{app:cost} gives the component-wise audit and the 77.3 GB allocated-memory peak, and Appendix~\ref{app:construction_ledger} unit-tests the analytical FLOP ledger against the launched tile shapes and verifies that the shared one-store implementation is not silently duplicating checkpoints (five duplicated 70B INT8 stores would need roughly 353 GB versus the observed 77.3 GB).

\section{Discussion and Conclusion}
\label{sec:conclusion}

The results support a conditional conclusion. When a frozen model exhibits out-of-sample spectral--task separability, calibrated spectral descriptors can guide task-conditioned, hardware-valid masks that retain more capability than a universal mask at the same active-FLOP budget. Compiling those masks converts analytical savings into measured latency gains, while confidence fallback exposes an explicit accuracy--efficiency operating curve. Neither routing alone nor mask multiplicity alone explains the result; spectral scoring and task-conditioned structure contribute complementary gains.

The same evidence also identifies where the method should not be used. Qwen2.5-1.5B remains near chance under the pilot, so further calibration is not justified. Although the Llama-3-8B results demonstrate transfer beyond 70B, the current study does not disentangle the effects of scale, architecture, training corpus, and model family. Likewise, the salience and ablation analyses identify reasoning-sensitive modules only in a predictive intervention sense, not as evidence of a unique reasoning circuit; stronger causal claims would require activation patching or causal mediation.

Within its supported regime, \tasp\ reduces active decoding FLOPs by 43\% while retaining $97.7{\pm}0.2\%$ in the BF16 benchmark harness. In the corrected single-A100 runtime with INT8 weight residency and BF16 compute, the sparse path achieves $97.3{\pm}0.2\%$ retention relative to dense BF16 with a $1.44\times$ decode speedup. More broadly, the work reframes spectral pruning as a deployment decision: establish separability, construct dependency-valid masks, and evaluate the resulting quality, latency, and amortization trade-offs.

Several extensions follow naturally. The applicability rule should be validated prospectively across broader model families; a first test on six unseen checkpoints (Appendix~\ref{app:prospective}) shows the frozen rule to be conservative, achieving $100\%$ precision, $75\%$ recall, and perfect monotone ordering (Spearman $\rho=1.00$ on six points) between pilot AUROC and downstream retention. Future work may also investigate activation patching, compiler-aware mask selection within the Rashomon set, and combinations of routed structured sparsity with quantization or low-rank approximation, each requiring end-to-end latency and retention evaluation. Supplementary visualizations of the main ablations and retention results are provided in Appendix~\ref{app:supplementary_figures}.
\section*{Limitations}

\tasp\ requires task definitions that remain stable long enough to amortize calibration. Its 136-GPU-hour 70B pipeline is substantially more expensive than the compared post-training baselines, and the retrospective pilot thresholds require prospective validation on more models. The current candidates target decoder-only Transformers with grouped-query attention and SwiGLU feed-forward blocks; encoder--decoder, state-space, mixture-of-experts, and multimodal architectures require new structural units and compatibility rules. The router can also be overconfident under task or language shift, so aggregate retention on the present English-dominated suite is not evidence of safety for unseen domains. Finally, the measured latency is specific to one A100 runtime, sequence shape, batch size, quantization scheme, and compiler. Analytical FLOP reductions should not be extrapolated to other systems without new measurements.

\section*{Ethics Statement}

Structured sparsity can reduce serving cost and thereby widen access to language models, but pruning may unevenly damage languages, domains, or demographic groups that are underrepresented in calibration. Deployment should therefore report task- and subgroup-specific quality rather than relying only on macro retention, and safety-critical traffic should use conservative fallback thresholds. The offline calibration cost also carries an environmental burden that is worthwhile only when the serving volume reaches the stated break-even regime. No human-subject data are introduced by the method; all benchmark and model use should continue to follow the licenses and intended-use restrictions of the underlying resources.

\bibliography{references}

\appendix

\section{Extended Related Work}
\label{app:related}

The main text positions \tasp\ at the intersection of structured pruning, spectral analysis, and conditional computation. This appendix makes the boundaries more explicit. LLM-Pruner removes dependency-aware structures using gradient information and may recover quality with LoRA; SliceGPT reduces width through aligned row and column removal; Sheared-LLaMA combines structured pruning with continued training; DISP-LLM relaxes structural constraints across embedding dimensions; and ShortGPT removes entire Transformer blocks \citep{ma2023llmpruner,ashkboos2024slicegpt,xia2024shearedllama,gao2024displlm,men2025shortgpt}. These methods differ in granularity and recovery cost, but each primarily produces one compressed architecture for a chosen operating point. \tasp\ instead asks whether several task-conditioned architectures can share one packed weight store and be selected without changing masks inside a generation.

Classical Random Matrix Theory characterizes spectra produced by high-dimensional random systems \citep{wigner1958distribution,marchenko1967distribution}. Empirical neural-network spectra often exhibit heavy tails rather than matching these reference laws, motivating descriptors such as tail exponents and effective rank \citep{martin2018implicit,martin2021implicit}. AlphaPruning uses heavy-tail metrics to assign layerwise sparsity \citep{lu2024alphapruning}. The important difference in \tasp\ is statistical rather than terminological: spectral descriptors are accepted only when they predict task-specific structural-ablation effects on out-of-fold modules, and the final claim is tested again on untouched modules and prompts.

Conditional computation usually changes the executed path through architectural design. Mixture-of-experts models route tokens among expert blocks \citep{shazeer2017outrageously,lepikhin2020gshard}; Flextron and LLaMaFlex expose elastic subnetworks at different budgets \citep{cai2024flextron,xu2024llamaflex}; and DeeBERT and CALM stop at intermediate depths when sufficient confidence is reached \citep{xin2020deebert,schuster2021confident}. \tasp\ does not introduce expert layers, modify token-wise computation, or exit early. It derives a small family of static, post-training subnetworks, routes once per turn, and keeps the selected mask fixed to preserve cache consistency.

\section{End-to-End Procedure}
\label{app:procedure}

Table~\ref{tab:procedure} expands the transitions in Figure~\ref{fig:pipeline}. The ordering is consequential: applicability is evaluated before the expensive portion of labeling, and compilation occurs before the router is timed.

\begin{table*}[t]
\centering
\small
\begin{tabularx}{\textwidth}{@{}p{.14\textwidth}p{.26\textwidth}X@{}}
\toprule
Stage & Inputs and split & Operation and output \\
\midrule
Candidate formation & Frozen checkpoint and architecture metadata & Slice every query head and each 1,024-channel SwiGLU group; record GQA and gate/up/down dependencies. \\
Streaming profile & One BF16 layer or matrix group at a time & Compute spectral descriptors without placing the full BF16 checkpoint on the GPU. \\
Pilot labels & Validation modules; 16 sequences per module--task pair & Measure structural-ablation targets and obtain out-of-fold predictions through nested, stratified module CV. \\
Applicability gate & Four diagnostic-family AUROCs and hierarchical bootstrap & Stop if mean $<0.62$, any family $<0.58$, or the mean 95\% LCB $<0.60$; otherwise finish label collection. \\
Final calibration & Train/validation modules and disjoint untouched test & Fit task-specific ridge predictors, select regularization on validation modules, and evaluate once on unseen modules and prompts. \\
Mask construction & Scores, active-FLOP budget, dependency graph & Greedily select by score per active FLOP, close GQA and SwiGLU dependencies, and emit one mask per task family. \\
Compilation & Five masks and shared INT8 backing store & Pack retained slices into contiguous tensors and build one static launch plan per task mask. \\
Routing & Labeled task prompts and held-out calibration prompts & Train the three-layer router, temperature-scale its probabilities, and choose the confidence threshold. \\
Inference & User prompt, compiled plans, and shared weights & Select a sparse plan or dense fallback once per turn; use the same plan for prefill and all decode steps. \\
\bottomrule
\end{tabularx}
\caption{The complete offline and online procedure. Only routing and dispatch are online.}
\label{tab:procedure}
\end{table*}

For Llama-3-70B, 80 layers times 92 candidates per layer gives 7,360 structured candidates. Stratification yields 5,152 training, 1,104 validation, and 1,104 untouched-test modules. The pilot is the first half of validation labeling, or $1{,}104\times5\times16=88{,}320$ prompt--module evaluations. Training and full validation use 32 sequences per module--task pair, and untouched testing uses 16, giving
\begin{equation}
 5(5{,}152+1{,}104)32+5(1{,}104)16=1{,}089{,}280
\end{equation}
evaluations in the complete pipeline.

\section{Detailed Pipeline Views}
\label{app:detailed_pipeline}

Figure~\ref{fig:pipeline} merges the complete decision path into the main paper, as requested in the meta-review. The original detailed diagrams are retained here because they expose the feature, dependency, and execution interfaces at a larger scale rather than introducing additional method variants.

\begin{figure*}[t]
\centering
\includegraphics[width=\textwidth]{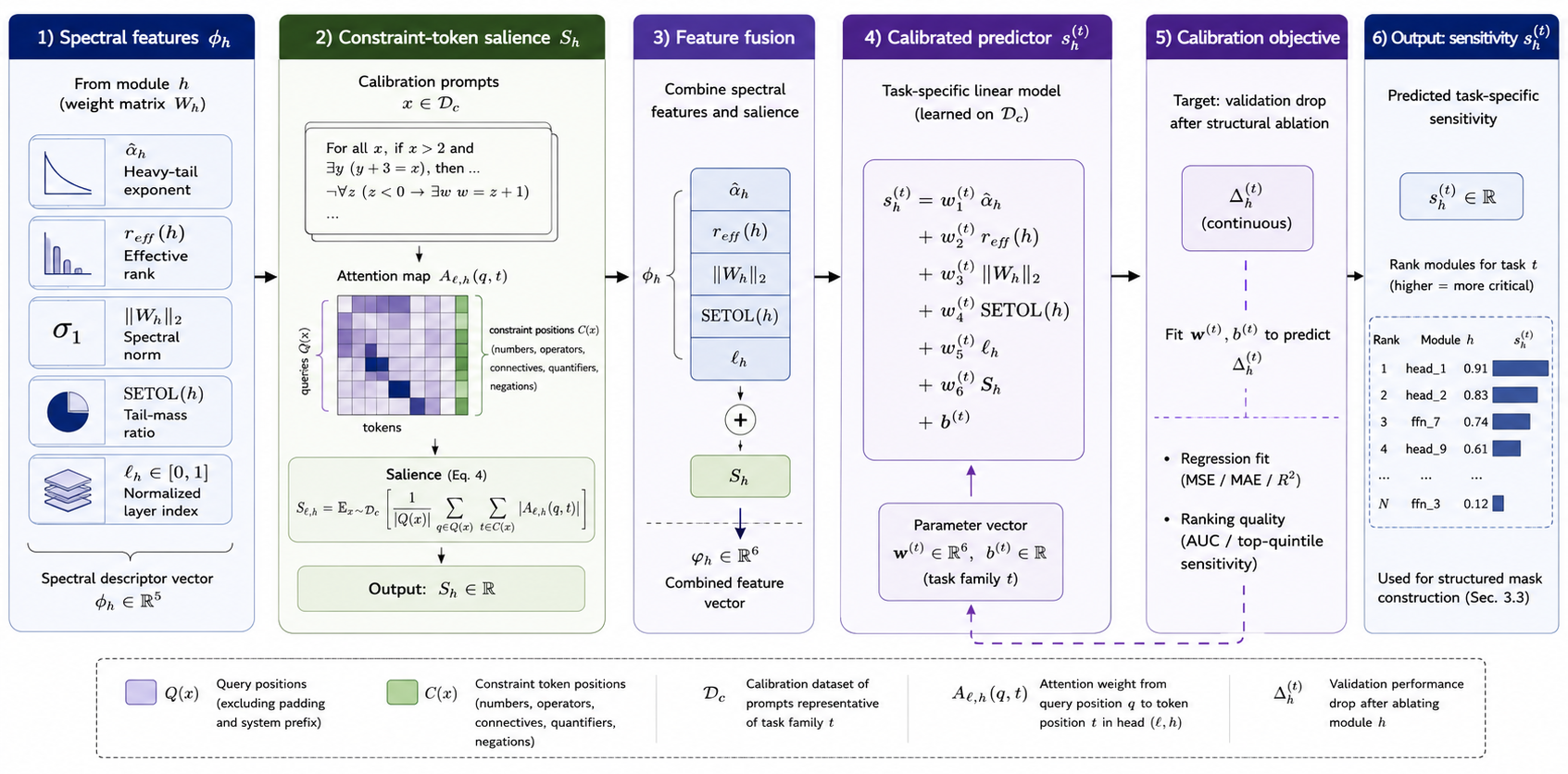}
\caption{Calibrated sensitivity estimation in \tasp. Module-level spectral descriptors and constraint-token salience predict task-specific degradation under structural ablation. Normalization, ridge selection, and evaluation follow the module-disjoint protocol described in Section~\ref{sec:method}.}
\label{fig:calibrated_sensitivity}
\end{figure*}

\begin{figure*}[t]
\centering
\includegraphics[width=\textwidth]{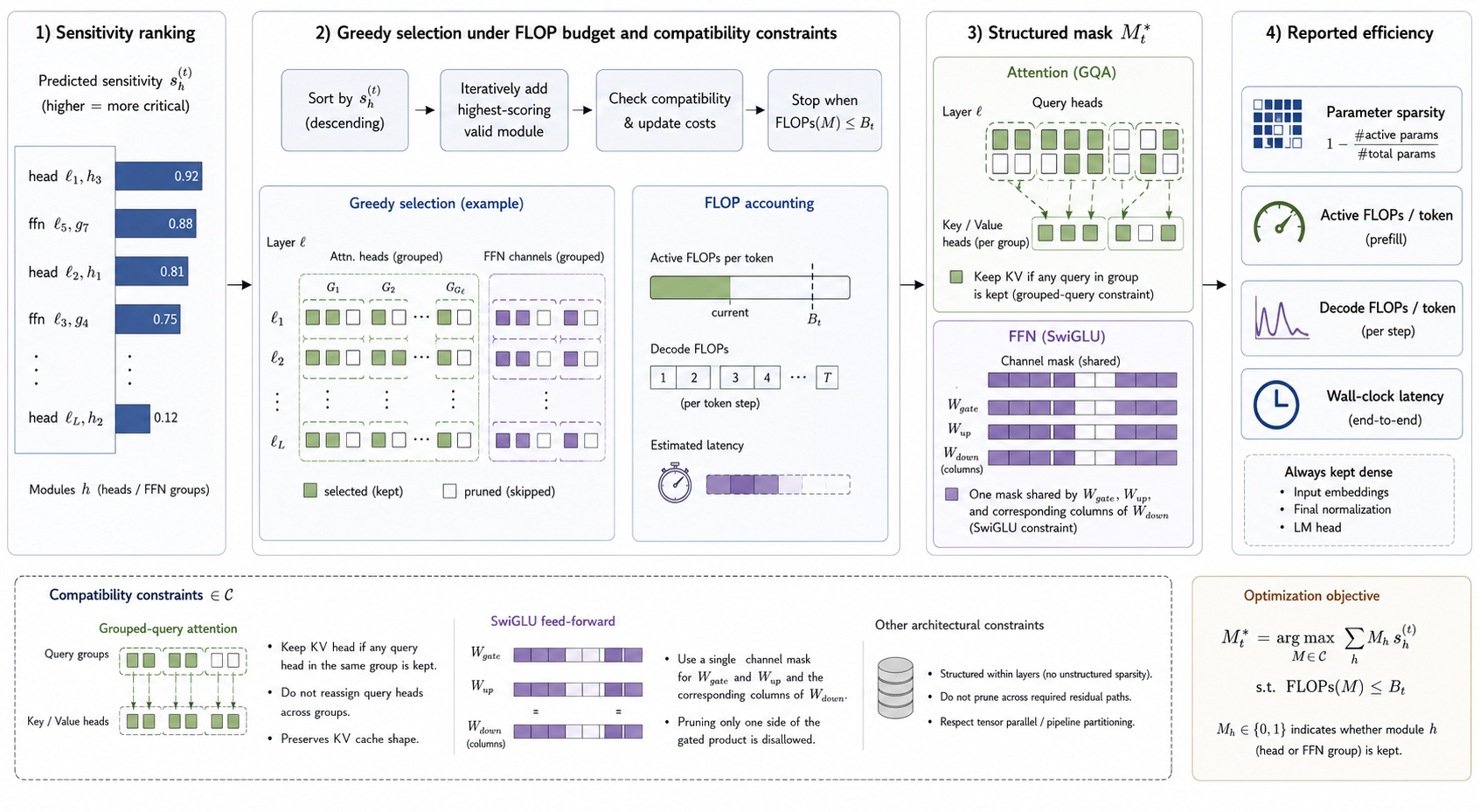}
\caption{Dependency-closed mask construction. Candidate modules are ranked by predicted sensitivity per active FLOP; GQA and SwiGLU dependencies are closed before a candidate is admitted, producing a hardware-valid static subnetwork at the target budget.}
\label{fig:structured_mask_construction}
\end{figure*}

\begin{figure*}[t]
\centering
\includegraphics[width=\textwidth]{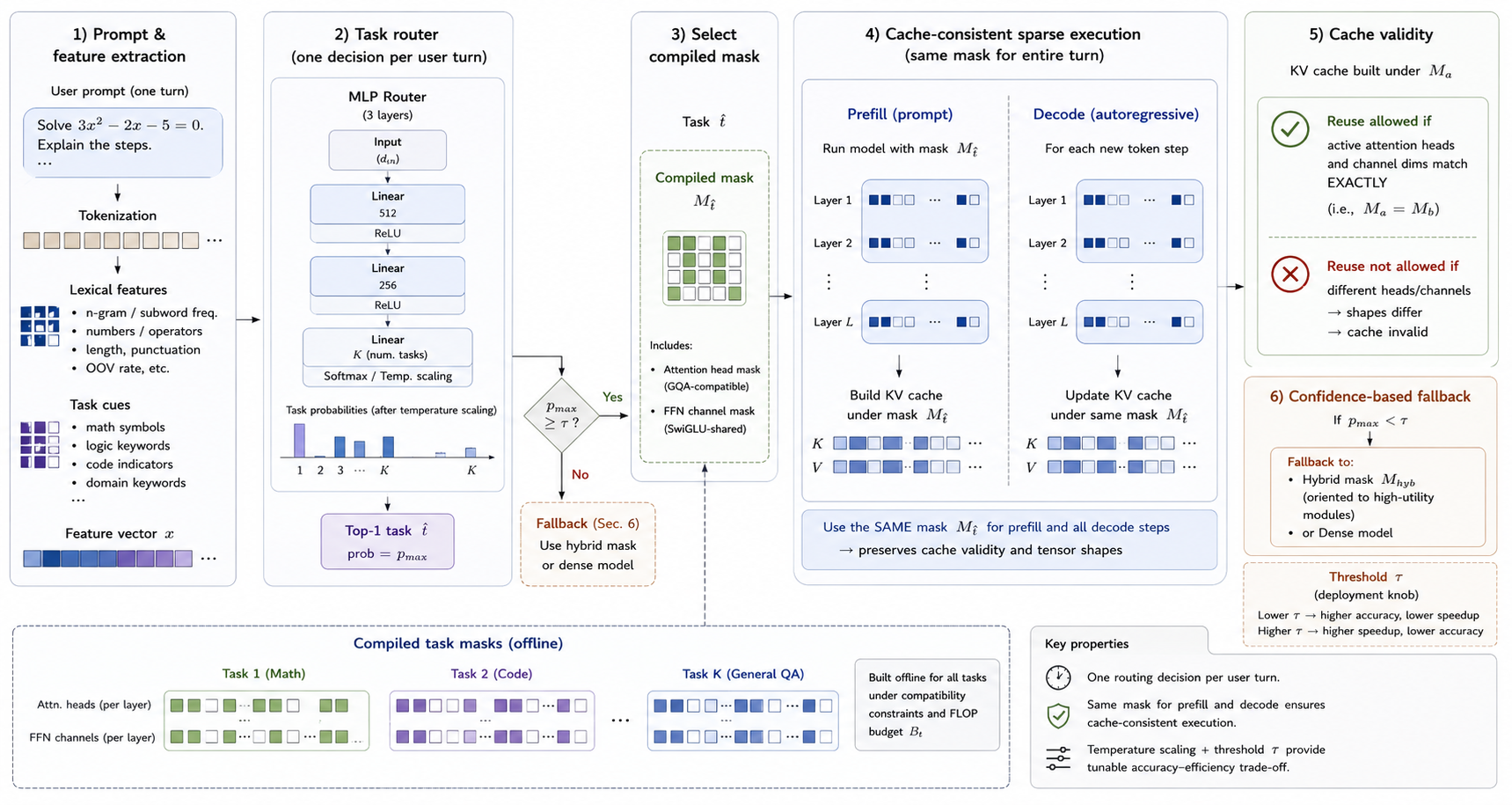}
\caption{Inference-time routing and cache-consistent execution. One mask is chosen per user turn and held fixed through prefill and decoding. Temperature-scaled low-confidence routes use the dense INT8 path, and KV caches are not reused across incompatible masks.}
\label{fig:routing_cache_consistent_execution}
\end{figure*}

Figure~\ref{fig:calibrated_sensitivity} illustrates how TASP converts structural descriptors into reliable sensitivity estimates for unseen modules, while Figure~\ref{fig:structured_mask_construction} summarizes the subsequent transformation of these predictions into dependency-valid static subnetworks under the target FLOP budget. Figure~\ref{fig:routing_cache_consistent_execution} further shows the inference-time routing mechanism that selects precompiled masks while preserving cache-consistent execution.

\section{Full Benchmark Results}
\label{app:benchmarks}

Table~\ref{tab:full_benchmarks} preserves the benchmark-level comparison summarized by relative retention in the main text. The displayed task columns are representative; the final retention column also includes MT-Bench, TriviaQA, MBPP, and both WMT14 directions. The table reports the BF16 scientific harness rather than the separately quantized timing runtime.

\begin{table*}[t]
\centering
\scriptsize
\resizebox{\textwidth}{!}{%
\begin{tabular}{lccccccccc}
\toprule
Method & FLOP red. & MMLU & GSM8K & MGSM & AlpacaEval & NQ-Open & HumanEval & WMT14 avg. & Full retention \\
\midrule
Dense BF16 & 0\% & $79.2$ & $87.5$ & $74.3$ & $82.1$ & $68.4$ & $72.6$ & $28.3$ & $100.0$ \\
\midrule
Magnitude & 43\% & $64.1{\pm}0.5$ & $58.3{\pm}0.7$ & $52.7{\pm}0.6$ & $67.8{\pm}0.5$ & $59.2{\pm}0.4$ & $43.1{\pm}0.8$ & $21.4{\pm}0.3$ & $75.6{\pm}0.5$ \\
SparseGPT & 43\% & $72.3{\pm}0.4$ & $75.4{\pm}0.6$ & $66.8{\pm}0.4$ & $74.2{\pm}0.4$ & $63.1{\pm}0.3$ & $58.7{\pm}0.6$ & $24.8{\pm}0.2$ & $88.5{\pm}0.4$ \\
Wanda & 43\% & $73.8{\pm}0.3$ & $77.1{\pm}0.5$ & $68.2{\pm}0.4$ & $75.6{\pm}0.4$ & $64.3{\pm}0.3$ & $61.2{\pm}0.5$ & $25.6{\pm}0.2$ & $90.7{\pm}0.3$ \\
Gradient & 43\% & $74.2{\pm}0.3$ & $78.3{\pm}0.4$ & $69.1{\pm}0.4$ & $76.1{\pm}0.3$ & $64.8{\pm}0.3$ & $62.3{\pm}0.5$ & $26.1{\pm}0.2$ & $91.7{\pm}0.3$ \\
AlphaPruning & 43\% & $75.0{\pm}0.3$ & $79.4{\pm}0.5$ & $69.7{\pm}0.4$ & $76.9{\pm}0.3$ & $65.4{\pm}0.3$ & $63.8{\pm}0.5$ & $26.4{\pm}0.2$ & $92.9{\pm}0.3$ \\
LLM-Pruner & 43\% & $74.9{\pm}0.4$ & $79.0{\pm}0.5$ & $69.4{\pm}0.4$ & $76.7{\pm}0.4$ & $65.0{\pm}0.3$ & $63.3{\pm}0.5$ & $26.2{\pm}0.2$ & $92.3{\pm}0.3$ \\
SliceGPT & 43\% & $75.4{\pm}0.3$ & $79.6{\pm}0.5$ & $69.9{\pm}0.4$ & $77.2{\pm}0.3$ & $65.8{\pm}0.3$ & $64.2{\pm}0.5$ & $26.5{\pm}0.2$ & $93.2{\pm}0.3$ \\
ShortGPT & 43\% & $74.6{\pm}0.4$ & $78.7{\pm}0.5$ & $69.0{\pm}0.4$ & $76.5{\pm}0.4$ & $64.9{\pm}0.3$ & $62.9{\pm}0.5$ & $26.0{\pm}0.2$ & $91.9{\pm}0.4$ \\
LLaMaFlex & 43\% & $75.8{\pm}0.3$ & $80.2{\pm}0.5$ & $70.4{\pm}0.3$ & $77.9{\pm}0.3$ & $66.1{\pm}0.2$ & $64.8{\pm}0.4$ & $26.9{\pm}0.2$ & $94.0{\pm}0.3$ \\
Agent-guided & 43\% & $74.1{\pm}0.4$ & $79.1{\pm}0.6$ & $68.7{\pm}0.4$ & $76.8{\pm}0.4$ & $65.2{\pm}0.3$ & $61.9{\pm}0.5$ & $25.8{\pm}0.2$ & $91.8{\pm}0.4$ \\
\midrule
\tasp-Single & 43\% & $75.1{\pm}0.3$ & $79.8{\pm}0.4$ & $70.4{\pm}0.3$ & $77.3{\pm}0.3$ & $65.9{\pm}0.2$ & $63.4{\pm}0.4$ & $26.4{\pm}0.2$ & $93.3{\pm}0.3$ \\
\tasp-\mom & 43\% & $77.9{\pm}0.3$ & $85.2{\pm}0.4$ & $72.8{\pm}0.3$ & $80.4{\pm}0.3$ & $67.2{\pm}0.2$ & $69.8{\pm}0.5$ & $27.6{\pm}0.2$ & $97.7{\pm}0.2$ \\
\tasp-Oracle & 43\% & $78.4{\pm}0.2$ & $86.1{\pm}0.3$ & $73.5{\pm}0.3$ & $81.3{\pm}0.2$ & $67.8{\pm}0.2$ & $70.9{\pm}0.4$ & $28.1{\pm}0.2$ & $98.7{\pm}0.2$ \\
\bottomrule
\end{tabular}}
\caption{Llama-3-70B quality at matched active-FLOP reduction. The reasoning column is original four-choice five-shot MMLU \citep{hendrycks2021mmlu}, corrected from the submitted ``MMLU-Pro'' label; the score is unchanged and a genuine MMLU-Pro run is reported separately in Appendix~\ref{app:mmlupro}. For \tasp\ variants, uncertainty is SD across five independent calibration/mask runs. For deterministic baselines, it is one paired-bootstrap SE across evaluation instances; these intervals should not be interpreted as variation from five independently selected baseline artifacts.}
\label{tab:full_benchmarks}
\end{table*}

\section{Accuracy--FLOP Pareto Frontier}
\label{app:pareto}

\begin{figure*}[t]
\centering
\includegraphics[width=.85\textwidth]{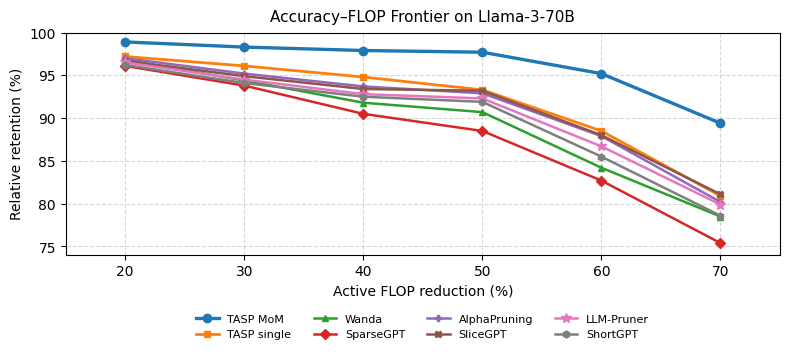}
\caption{Accuracy--FLOP frontier on Llama-3-70B. Relative retention is plotted against active decoding-FLOP reduction for \tasp\ and representative pruning baselines. The matched 43\% operating point is the only point for which all structured systems are additionally compiled and timed under the deployment runtime.}
\label{fig:pareto}
\end{figure*}

The accuracy--FLOP sweep from 20\% to 70\% reduction shows the same qualitative ordering as the matched operating point: \tasp-\mom\ remains above the global mask over the 30--60\% range, and the learned-to-oracle gap widens at the most aggressive sparsity. We retain the matched 43\% point as the primary comparison because it is the point for which every structured baseline was compiled and timed under the same runtime. Figure~\ref{fig:pareto} retains the complete frontier from the original analysis rather than reducing it to the single matched point.

\section{Applicability and Disjoint Diagnostics}
\label{app:diagnostics}

\begin{figure*}[t]
\centering
\begin{tikzpicture}
\begin{axis}[
  width=.82\textwidth,
  height=.31\textwidth,
  ybar,
  bar width=8pt,
  ymin=.45,
  ymax=.90,
  ylabel={Out-of-fold AUROC},
  symbolic x coords={Reasoning,Retrieval,Code,Translation,Mean},
  xtick=data,
  x tick label style={font=\small},
  legend style={font=\small, at={(.5,1.03)}, anchor=south, legend columns=3},
  ymajorgrids=true,
  grid style={dotted}
]
\addplot coordinates {(Reasoning,.53) (Retrieval,.52) (Code,.55) (Translation,.51) (Mean,.53)};
\addplot coordinates {(Reasoning,.69) (Retrieval,.65) (Code,.67) (Translation,.63) (Mean,.66)};
\addplot coordinates {(Reasoning,.82) (Retrieval,.76) (Code,.79) (Translation,.74) (Mean,.78)};
\legend{Qwen2.5-1.5B,Llama-3-8B,Llama-3-70B}
\end{axis}
\end{tikzpicture}
\caption{Predeployment spectral--task separability. The figure reports only out-of-fold predictions from the pilot protocol.}
\label{fig:diagnostic}
\label{fig:auc}
\end{figure*}
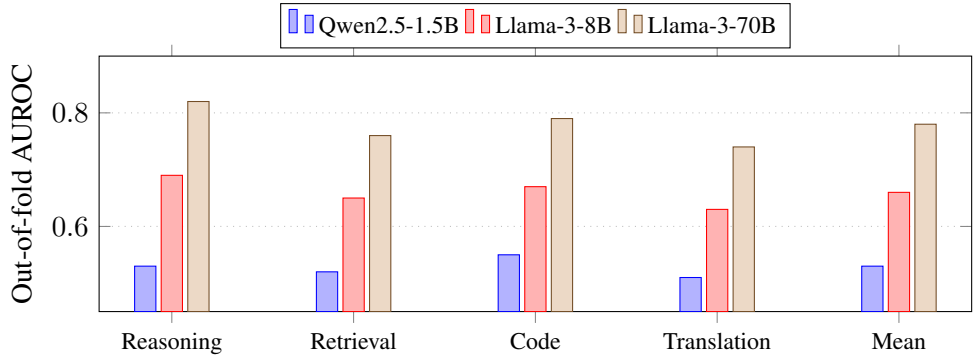

Figure~\ref{fig:diagnostic} visualizes the gate values for all three evaluated checkpoints. Llama-3-8B occupies an intermediate regime between the near-chance Qwen pilot and the stronger 70B separation. This pattern is compatible with several explanations, including scale, architecture, family, and training data; three checkpoints cannot distinguish them. Table~\ref{tab:test_auc} reports the out-of-fold pilot and final disjoint evaluation AUROC results, showing the generalization of the sensitivity predictor across task families. Additional transfer and compression controls are reported in Appendix~\ref{app:transfer}. Table~\ref{tab:8b} reports direct Llama-3-8B deployment results at the matched 43\% active-FLOP reduction, showing task-family retention and full-suite performance of TASP-MOM. 

\begin{table}[t]
\centering
\scriptsize
\setlength{\tabcolsep}{3pt}
\begin{tabular}{lccr}
\toprule
Task & Pilot AUROC & Test AUROC & Change \\
\midrule
Reasoning & 0.82 & 0.80 & $-0.02$ \\
Retrieval & 0.76 & 0.72 & $-0.04$ \\
Code & 0.79 & 0.76 & $-0.03$ \\
Translation & 0.74 & 0.73 & $-0.01$ \\
Mean & 0.78 & 0.75 & $-0.03$ \\
\bottomrule
\end{tabular}
\caption{Llama-3-70B out-of-fold pilot performance and final evaluation on untouched modules and disjoint prompts.}
\label{tab:test_auc}
\end{table}

\begin{table}[t]
\centering
\small
\begin{tabular}{lc}
\toprule
Llama-3-8B task family & \tasp-\mom\ retention \\
\midrule
Reasoning & $95.8{\pm}0.4$ \\
Generation & $97.1{\pm}0.3$ \\
Retrieval & $96.8{\pm}0.3$ \\
Code & $95.6{\pm}0.5$ \\
Translation & $96.7{\pm}0.3$ \\
Full-suite mean & $96.4{\pm}0.3$ \\
\bottomrule
\end{tabular}
\caption{Direct Llama-3-8B deployment at 43\% active-FLOP reduction. The full-suite mean is computed from eleven benchmark-level ratios; family rows are displayed for readability.}
\label{tab:8b}
\end{table}

As an additional descriptive check, the lowest 20\% of calibrated tail-exponent ranks account for 63\% of measured 70B reasoning sensitivity, compared with 28\% for a layer-matched random control. The stronger intervention test is the unseen-module removal comparison in Table~\ref{tab:causal_test}.

\begin{table}[t]
\centering
\small
\begin{tabular}{lc}
\toprule
Selection rule & Mean degradation \\
\midrule
\tasp\ least-important groups & $-2.6{\pm}0.3$ \\
Wanda least-important groups & $-4.8{\pm}0.3$ \\
Layer-matched random groups & $-7.4{\pm}0.4$ \\
\bottomrule
\end{tabular}
\caption{Performance-point degradation after removing groups on untouched 70B modules and disjoint prompts. Values are mean $\pm$ SD over five runs; smaller absolute degradation is better.}
\label{tab:causal_test}
\end{table}

\section{Exploratory Taskwise Ablation Diagnostic}
\label{app:causal}

\begin{figure}[t]
\centering
\includegraphics[width=\columnwidth]{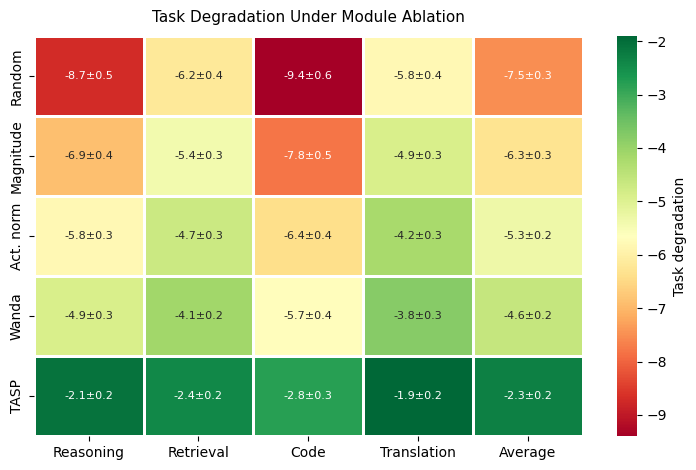}
\caption{Taskwise visualization of the exploratory diagnostic in Table~\ref{tab:causal_ablation}. The figure is retained to show the family-level pattern, while the main evidence is the disjoint-module and disjoint-prompt analysis.}
\label{fig:task_degradation_heatmap}
\end{figure}

The submitted paper also contained the taskwise diagnostic in Table~\ref{tab:causal_ablation}. It is retained here because it shows how the ranking behaves within individual task families, but its scope is now explicit: the analysis uses the original calibration/validation module protocol and is exploratory. It does not replace the confirmatory module- and prompt-disjoint result in Table~\ref{tab:causal_test}. The two tables therefore answer different questions rather than providing competing estimates of the same quantity. Exploratory taskwise trends are visualized in Figure~\ref{fig:task_degradation_heatmap}, while the primary module- and prompt-disjoint evaluation is retained as the main assessment.

\begin{table*}[t]
\centering
\scriptsize
\begin{tabular}{lccccc}
\toprule
Rule & Reasoning & Retrieval & Code & Translation & Average \\
\midrule
Random & $-8.7{\pm}0.5$ & $-6.2{\pm}0.4$ & $-9.4{\pm}0.6$ & $-5.8{\pm}0.4$ & $-7.5{\pm}0.3$ \\
Magnitude & $-6.9{\pm}0.4$ & $-5.4{\pm}0.3$ & $-7.8{\pm}0.5$ & $-4.9{\pm}0.3$ & $-6.3{\pm}0.3$ \\
Activation norm & $-5.8{\pm}0.3$ & $-4.7{\pm}0.3$ & $-6.4{\pm}0.4$ & $-4.2{\pm}0.3$ & $-5.3{\pm}0.2$ \\
Wanda & $-4.9{\pm}0.3$ & $-4.1{\pm}0.2$ & $-5.7{\pm}0.4$ & $-3.8{\pm}0.3$ & $-4.6{\pm}0.2$ \\
\tasp & $-2.1{\pm}0.2$ & $-2.4{\pm}0.2$ & $-2.8{\pm}0.3$ & $-1.9{\pm}0.2$ & $-2.3{\pm}0.2$ \\
\bottomrule
\end{tabular}
\caption{Exploratory validation-module degradation after removing the groups ranked least important by each rule at matched structured sparsity. Smaller absolute degradation is better. Confirmatory claims use the disjoint protocol in Table~\ref{tab:causal_test}.}
\label{tab:causal_ablation}
\end{table*}

\section{Routing, Calibration, and Task-Switching Robustness}
\label{app:robustness}

Table~\ref{tab:threshold_sweep} reports the natural deployment curve. Below-threshold requests use dense INT8; the hybrid path is disabled. The decode values are mixture-implied from the separately measured dense and sparse paths,
\begin{equation}
 L(\tau)=u(\tau)L_{\mathrm{sparse}}+[1-u(\tau)]L_{\mathrm{dense}},
\end{equation}
where $u(\tau)$ is sparse-path usage. Only the all-sparse endpoint is a directly timed sparse-kernel result.

\begin{table}[t]
\centering
\scriptsize
\setlength{\tabcolsep}{2.5pt}
\resizebox{\columnwidth}{!}{%
\begin{tabular}{ccccc}
\toprule
$\tau$ & Ret. vs. INT8 & Sparse use & ms/token & Speedup \\
\midrule
0.50 & $97.7{\pm}0.2$ & $100{\pm}0\%$ & 31.3 & $1.44\times$ \\
0.70 & $98.4{\pm}0.2$ & $91{\pm}2\%$ & 32.6 & $1.39\times$ \\
0.85 & $99.3{\pm}0.1$ & $76{\pm}3\%$ & 34.6 & $1.30\times$ \\
0.95 & $99.7{\pm}0.1$ & $52{\pm}3\%$ & 38.0 & $1.19\times$ \\
\bottomrule
\end{tabular}
}
\caption{Natural dense-fallback operating curve. Latency is mixture-implied from measured path latencies.}
\label{tab:threshold_sweep}
\label{tab:calibration}
\end{table}

To isolate route correctness from natural confidence, Table~\ref{tab:route_corruption} replaces a fixed fraction of predictions with confusion-matrix-matched incorrect routes. The 100\% row assigns constructed above-threshold confidence to every correct route; its 0\% fallback rate is therefore not the confidence distribution in Table~\ref{tab:threshold_sweep}. Corrupted rows draw confidence from the empirical error-confidence distribution.

\begin{table}[t]
\centering
\scriptsize
\resizebox{\columnwidth}{!}{%
\begin{tabular}{cccc}
\toprule
Top-1 accuracy & No fallback & With $\tau=0.85$ & Fallback rate \\
\midrule
100\% & $98.7{\pm}0.2$ & $98.7{\pm}0.2$ & 0\% \\
90\% & $97.8{\pm}0.2$ & $98.4{\pm}0.2$ & 12\% \\
80\% & $96.6{\pm}0.3$ & $98.0{\pm}0.2$ & 24\% \\
70\% & $95.2{\pm}0.4$ & $97.5{\pm}0.3$ & 37\% \\
\bottomrule
\end{tabular}
}
\caption{Controlled routing-corruption study in the BF16 benchmark harness.}
\label{tab:route_corruption}
\end{table}

For the limited-label study, a fraction of complete training module--prompt pairs is sampled jointly without replacement and stratified by task family, layer, and module type. Validation modules, untouched-test modules, validation prompts, and test prompts remain fixed. Thus, the 50\% condition uses exactly half of the labels rather than independently retaining half the modules and half the prompts. Table~\ref{tab:label_fraction} reports the effect of calibration-label volume on held-out AUROC and TASP-MOM retention, demonstrating sensitivity to available supervision.

\begin{table}[t]
\centering
\scriptsize
\setlength{\tabcolsep}{3pt}
\begin{tabular}{ccc}
\toprule
Labels & Held-out AUROC & \tasp-\mom\ ret. \\
\midrule
25\% & $0.66{\pm}0.03$ & $95.7{\pm}0.4$ \\
50\% & $0.70{\pm}0.02$ & $96.7{\pm}0.3$ \\
75\% & $0.73{\pm}0.02$ & $97.2{\pm}0.3$ \\
100\% & $0.75{\pm}0.02$ & $97.7{\pm}0.2$ \\
\bottomrule
\end{tabular}
\caption{Sensitivity to calibration-label volume. Values are mean $\pm$ SD over five stratified label subsets.}
\label{tab:label_fraction}
\end{table}

The one-mask-per-turn rule does not mean one mask per conversation. Table~\ref{tab:task_switching} alternates task families across eight turns and reroutes after each user turn. Per-turn routing closes most of the gap to dense, whereas locking the first mask allows degradation to accumulate after the task changes.

\begin{table}[t]
\centering
\scriptsize
\resizebox{\columnwidth}{!}{%
\begin{tabular}{lccc}
\toprule
Scenario & Dense & First-turn locked & Per-turn routing \\
\midrule
Math $\rightarrow$ creative & $75.2$ & $71.8{\pm}0.4$ & $74.1{\pm}0.4$ \\
Code $\rightarrow$ translation & $68.4$ & $64.9{\pm}0.5$ & $67.1{\pm}0.5$ \\
Retrieval $\rightarrow$ reasoning & $72.8$ & $69.6{\pm}0.4$ & $71.9{\pm}0.3$ \\
Multi-task average & $72.1$ & $68.8{\pm}0.4$ & $71.0{\pm}0.4$ \\
\bottomrule
\end{tabular}
}
\caption{Eight-turn task-switching stress test. Uncertainty is SD across five runs.}
\label{tab:task_switching}
\label{tab:stress}
\end{table}

\begin{figure}[t]
\centering
\begin{tikzpicture}
\begin{axis}[
  width=.47\textwidth,
  height=.31\textwidth,
  xlabel={Fallback threshold $\tau$},
  ylabel={Percentage},
  xmin=.47,
  xmax=.98,
  ymin=45,
  ymax=102,
  xtick={.50,.70,.85,.95},
  ymajorgrids=true,
  grid style={dotted},
  legend style={font=\scriptsize, at={(.5,1.02)}, anchor=south, legend columns=2},
  tick label style={font=\scriptsize},
  label style={font=\small}
]
\addplot+[mark=*, thick] coordinates {(.50,97.7) (.70,98.4) (.85,99.3) (.95,99.7)};
\addplot+[mark=square*, thick] coordinates {(.50,100) (.70,91) (.85,76) (.95,52)};
\legend{Retention vs. INT8,Sparse-path usage}
\end{axis}
\end{tikzpicture}
\caption{Corrected router fallback trade-off. Higher thresholds recover dense-INT8 retention by sending more requests to the dense path. Decode latency is mixture-implied and reported numerically in Table~\ref{tab:threshold_sweep}.}
\label{fig:router_fallback_tradeoff}
\label{fig:router_curve}
\end{figure}
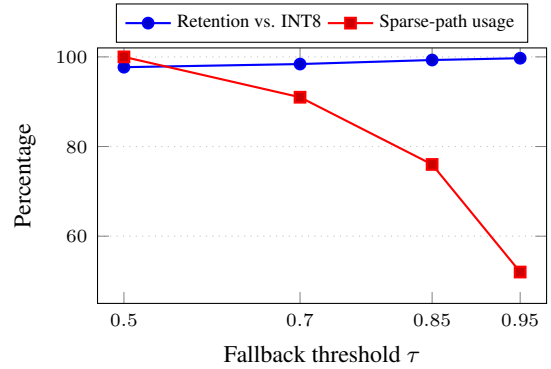

\begin{figure}[t]
\centering
\includegraphics[width=\columnwidth]{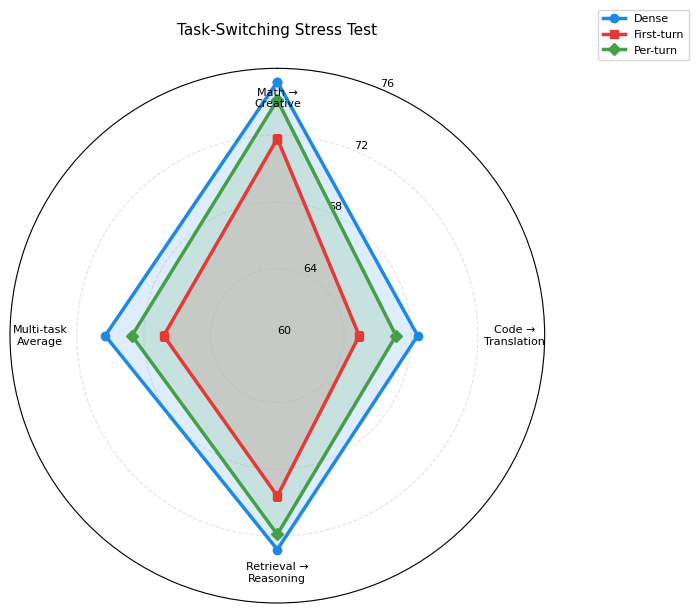}
\caption{Task-switching stress test across three representative domain transitions. Rerouting at each user turn substantially closes the gap created by locking a conversation to its first mask.}
\label{fig:task_switching_stress_test}
\end{figure}

Figure~\ref{fig:router_curve} visualizes the trade-off between router fallback threshold, retention, and sparse-path utilization, complementing the numerical analysis in Table~\ref{tab:calibration}.

Figure~\ref{fig:task_switching_stress_test} evaluates turn-level routing stability under task shifts, showing that per-turn rerouting preserves performance across changing user intents.

\section{Cross-Task Structure and Within-Task Multiplicity}
\label{app:masks}

Cross-task overlap measures specialization, whereas within-task overlap measures non-uniqueness. Table~\ref{tab:cross_overlap} shows that related task pairs share much more structure than reasoning and generation. Table~\ref{tab:within_overlap} then perturbs equal-score ties and the last 5\% of score-ranked selections while preserving the exact FLOP budget and structural constraints. Every task admits multiple near-equivalent masks.

\begin{table}[t]
\centering
\small
\begin{tabular}{lc}
\toprule
Task-mask pair & Jaccard overlap \\
\midrule
Reasoning / Code & $0.78{\pm}0.02$ \\
Generation / Translation & $0.81{\pm}0.02$ \\
Reasoning / Generation & $0.24{\pm}0.03$ \\
Retrieval / Translation & $0.52{\pm}0.03$ \\
Code / Translation & $0.43{\pm}0.03$ \\
\bottomrule
\end{tabular}
\caption{Selected cross-task mask overlaps across five calibration runs.}
\label{tab:cross_overlap}
\label{tab:overlap}
\end{table}

\begin{table*}[t]
\centering
\small
\begin{tabular}{lccc}
\toprule
Task & Across-run within-task Jaccard & Distinct masks within 0.5 points & Retention range \\
\midrule
Reasoning & $0.82{\pm}0.03$ & 13 & 97.0--97.3 \\
Retrieval & $0.78{\pm}0.04$ & 18 & 97.2--97.6 \\
Code & $0.81{\pm}0.03$ & 10 & 96.7--97.1 \\
Translation & $0.84{\pm}0.03$ & 16 & 97.9--98.2 \\
\bottomrule
\end{tabular}
\caption{Within-task multiplicity under score-tie and boundary perturbations. \tasp\ requires a high-retention, hardware-valid mask, not an identifiable unique circuit.}
\label{tab:within_overlap}
\end{table*}

This Rashomon structure can be operationally useful: once several masks satisfy the same quality and FLOP constraints, a compiler may choose among them according to device-specific matrix shapes. Evaluating that secondary compiler objective across hardware is left for future work.

\begin{figure}[t]
\centering
\includegraphics[width=\columnwidth]{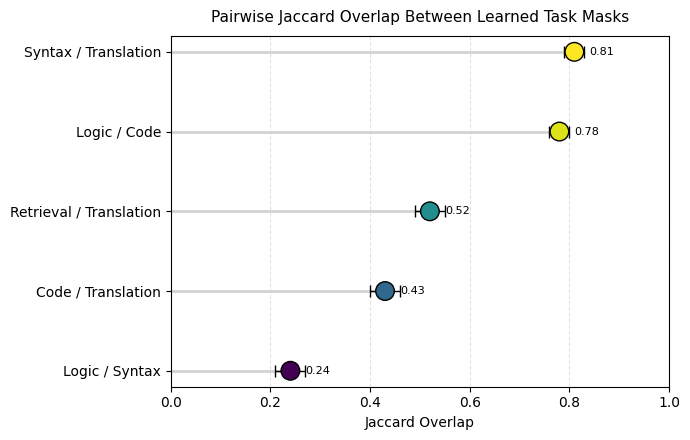}
\caption{Pairwise Jaccard overlap between learned task masks. Related task families share a larger stable core, whereas reasoning and generation rely on substantially more distinct retained structures.}
\label{fig:jaccard_overlap_masks}
\end{figure}

Figure~\ref{fig:jaccard_overlap_masks} visualizes task-mask overlap patterns by showing both shared structure across tasks and the existence of multiple valid sparse solutions.

\section{Offline Cost, Memory, and Amortization}
\label{app:cost}

Table~\ref{tab:offline_cost} uses a common accounting boundary: costs begin after the pretrained checkpoint is locally available and include method-native calibration or optimization, selection, INT8 packing, and compilation into the timed runtime. They exclude the shared checkpoint download and final benchmark evaluation. The pilot and test contain the same number of prompt--module evaluations, but the test is slower because it enforces cold module-plan reloads between task families, writes per-example outputs, and retains hierarchical-bootstrap traces. Table~\ref{tab:baseline_cost} compares the calibration cost and peak memory usage of TASP with representative structured pruning baselines under a common accounting protocol.

\begin{table*}[t]
\centering
\scriptsize
\resizebox{\textwidth}{!}{%
\begin{tabular}{lrrr}
\toprule
Offline component & Prompt--module evaluations & A100 GPU-hours & Peak allocated memory \\
\midrule
Layer-streamed BF16 spectral profiling & 0 & 5.6 & 11.8 GB \\
15\% pilot separability stage & 88,320 & 10.2 & 77.1 GB \\
Remaining train/validation labeling & 912,640 & 102.6 & 77.3 GB \\
Module- and prompt-disjoint test labeling & 88,320 & 11.7 & 77.3 GB \\
Predictor fitting and cross-validation & --- & 0.3 & 3.2 GB \\
FLOP-constrained mask construction & --- & 0.2 & 1.7 GB \\
Compilation of five task masks & --- & 4.8 & 77.3 GB \\
Router training and temperature scaling & --- & 0.6 & 5.4 GB \\
\midrule
Total \tasp & 1,089,280 & 136.0 & --- \\
\bottomrule
\end{tabular}
}
\caption{Complete Llama-3-70B offline cost audit. Ablation labels are conditioned on the deployment quantization because they are measured in the same INT8-weight/BF16-compute runtime.}
\label{tab:offline_cost}
\end{table*}

\begin{table}[t]
\centering
\small
\begin{tabular}{lcc}
\toprule
Method & A100 GPU-hours & Peak allocated \\
\midrule
Wanda & 7.4 & 76.5 GB \\
SparseGPT & 13.1 & 77.0 GB \\
SliceGPT & 10.3 & 76.8 GB \\
LLM-Pruner & 19.6 & 77.1 GB \\
\tasp & 136.0 & 77.3 GB \\
\bottomrule
\end{tabular}
\caption{Baseline calibration cost under the same accounting boundary. Baselines use 2,560 examples, except LLM-Pruner, which uses 25,600.}
\label{tab:baseline_cost}
\end{table}

The single-A100 measurement refers to allocated bytes, not to BF16 weight residency. The runtime has one packed weight backing store; the five masks are block-selection and launch metadata rather than duplicated checkpoints. Table~\ref{tab:memory} sums to the observed $\texttt{torch.cuda.max\_memory\_allocated}$ peak. Decimal GB are used; 77.3 GB is approximately 72.0 GiB. Reserved-memory and process-level peaks were not present in the audited timing export, so we do not substitute allocated memory for either measurement. 

\begin{table}[t]
\centering
\small
\begin{tabular}{lr}
\toprule
Live component & Allocation \\
\midrule
Packed INT8 weights & 70.60 GB \\
Scales and quantization metadata & 1.10 GB \\
Five mask plans and routing metadata & 0.54 GB \\
KV cache, activations, and workspace & 5.06 GB \\
\midrule
Peak allocated & 77.30 GB \\
\bottomrule
\end{tabular}
\caption{Observed live allocation during the complete 2,048-prefill/256-decode loop.}
\label{tab:memory}
\end{table}

The request-level amortization follows directly from Table~\ref{tab:systems}. Dense execution takes $1.814+256(0.0452)=13.385$ seconds, whereas the sparse path takes $1.326+256(0.0313)=9.339$ seconds. The 4.046-second difference yields $136\times3{,}600/4.046\approx121{,}000$ requests. Under $\tau=0.85$, 76\% sparse usage saves approximately 3.08 seconds per request and yields about 159,000 requests. These are workload-specific break-even points rather than universal efficiency claims.

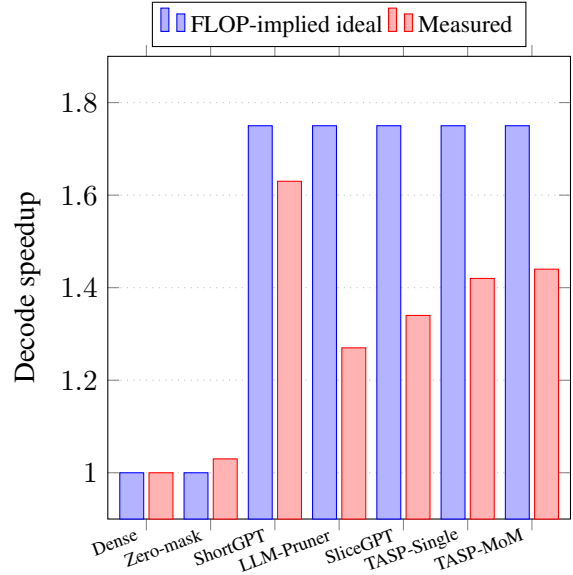
\begin{figure}[t]
\centering
\begin{tikzpicture}
\begin{axis}[
  width=\columnwidth,
  height=\columnwidth,
  ybar,
  bar width=9pt,
  ymin=0.9,
  ymax=1.9,
  ylabel={Decode speedup},
  symbolic x coords={Dense,Zero-mask,ShortGPT,LLM-Pruner,SliceGPT,TASP-Single,TASP-MoM},
  xtick=data,
  x tick label style={font=\scriptsize, rotate=20, anchor=east},
  ymajorgrids=true,
  grid style={dotted},
  legend style={font=\small, at={(.5,1.02)}, anchor=south, legend columns=2}
]
\addplot coordinates {(Dense,1.00) (Zero-mask,1.00) (ShortGPT,1.75) (LLM-Pruner,1.75) (SliceGPT,1.75) (TASP-Single,1.75) (TASP-MoM,1.75)};
\addplot coordinates {(Dense,1.00) (Zero-mask,1.03) (ShortGPT,1.63) (LLM-Pruner,1.27) (SliceGPT,1.34) (TASP-Single,1.42) (TASP-MoM,1.44)};
\legend{FLOP-implied ideal,Measured}
\end{axis}
\end{tikzpicture}
\caption{Corrected realized-versus-theoretical compute reduction. The 0.57-active-FLOP structured models have an idealized $1/0.57\approx1.75\times$ compute ratio, but measured speed depends on the compiled shapes. Applying the same selected \tasp\ masks as zeros in dense kernels launches approximately the dense workload and yields only $1.03\times$. This figure replaces the submitted visualization based on an incomplete profiler subtotal.}
\label{fig:realized_compute_reduction}
\end{figure}

Figure~\ref{fig:realized_compute_reduction} compares analytical FLOP savings with compiled runtime speedups, highlighting the gap between sparse mask quality and physically realized acceleration.

\section{Transfer and Compression Controls}
\label{app:transfer}

Direct Llama-3-8B deployment and cross-size transfer answer different questions. For transfer, we compute the calibrated ranking on Llama-3-8B, align normalized layer regions with Llama-3-70B, and construct a 70B mask without direct 70B ranking calibration. At the same 43\% active-FLOP reduction, that mask recovers 92\% of the performance obtained by direct 70B calibration. The largest gaps occur on GSM8K and HumanEval, where the identities of high-sensitivity heads differ most across sizes. The result indicates partial reuse within a family, not a replacement for target-model calibration.

A matched low-rank control replaces structural removal of each candidate by a rank-$k$ approximation at the same active-FLOP budget. Structured pruning is stronger for attention heads, by 1.9 retention points on GSM8K and 1.4 on HumanEval. Low-rank approximation is competitive for feed-forward matrices, remaining within 0.3 points on AlpacaEval and 0.5 on NQ-Open. Spectral descriptors may therefore guide either compression mode, but the preferred transformation depends on module type.

\section{Statistics, Seeds, and Reproducibility}
\label{app:statistics}

For \tasp, a seed controls calibration-prompt subsampling, ablation batch order, router initialization, equal-score mask tie-breaking, and evaluation ordering when supported by a benchmark. Each reported \tasp\ SD is computed across five independently generated calibration, mask, and router runs. Dense BF16 uses greedy decoding and is deterministic. After validation selects an artifact for a deterministic pruning baseline, its reported uncertainty is obtained by paired bootstrap over evaluation instances, not by presenting bootstrap draws as new pruning runs. Hierarchical bootstrap intervals for applicability resample modules within layer/type strata and prompts within task family.

Every latency row uses five independent timing blocks. A block contains 50 untimed warmups followed by 200 measured requests; CUDA events enclose the complete prefill or decode phase, and terminal synchronization is included. Router and dispatch time are charged once per request and amortized across generated tokens. Model loading, one-time packing, and compilation are outside steady-state timing. The method manifest records the source repository and commit, method-native search grid, selected validation setting, calibration data, quantization parameters, compiler configuration, and exact runtime boundary for each baseline.

Tables~\ref{tab:seed_retention}, \ref{tab:seed_ablation}, and \ref{tab:seed_tasks} expose the \tasp\ seed values underlying the main comparisons. Deterministic baseline rows are intentionally absent from these tables because their uncertainty has a different source.

\begin{table}[t]
\centering
\scriptsize
\resizebox{\columnwidth}{!}{%
\begin{tabular}{lcccccc}
\toprule
Variant & S1 & S2 & S3 & S4 & S5 & Mean $\pm$ SD \\
\midrule
\tasp-Single & 92.9 & 93.1 & 93.7 & 93.5 & 93.3 & $93.3{\pm}0.3$ \\
\tasp-\mom & 97.5 & 97.6 & 98.0 & 97.8 & 97.7 & $97.7{\pm}0.2$ \\
\tasp-Oracle & 98.5 & 98.6 & 99.0 & 98.8 & 98.7 & $98.7{\pm}0.2$ \\
\bottomrule
\end{tabular}
}
\caption{Seed-level Llama-3-70B retention in the BF16 benchmark harness.}
\label{tab:seed_retention}
\end{table}

\begin{table}[t]
\centering
\scriptsize
\setlength{\tabcolsep}{5pt}
\begin{tabular}{lcccccc}
\toprule
Variant & S1 & S2 & S3 & S4 & S5 & Mean $\pm$ SD \\
\midrule
Oracle router & 98.5 & 98.6 & 99.0 & 98.8 & 98.7 & $98.7{\pm}0.2$ \\
Learned router & 97.5 & 97.6 & 98.0 & 97.8 & 97.7 & $97.7{\pm}0.2$ \\
Without salience & 95.2 & 95.4 & 95.9 & 95.6 & 95.5 & $95.5{\pm}0.3$ \\
Without spectral features & 93.0 & 93.4 & 94.1 & 93.8 & 93.7 & $93.6{\pm}0.4$ \\
One global mask & 92.9 & 93.1 & 93.7 & 93.5 & 93.3 & $93.3{\pm}0.3$ \\
Layer-matched random & 85.0 & 85.4 & 86.2 & 86.0 & 85.5 & $85.6{\pm}0.5$ \\
\bottomrule
\end{tabular}
\caption{Seed-level ablation ladder corresponding to Table~\ref{tab:ablation}.}
\label{tab:seed_ablation}
\end{table}

\begin{table}[t]
\centering
\scriptsize
\setlength{\tabcolsep}{5pt}
\begin{tabular}{lcccccc}
\toprule
Metric & S1 & S2 & S3 & S4 & S5 & Mean $\pm$ SD \\
\midrule
MMLU & 77.5 & 77.8 & 78.3 & 78.0 & 77.9 & $77.9{\pm}0.3$ \\
GSM8K & 84.7 & 85.0 & 85.7 & 85.4 & 85.2 & $85.2{\pm}0.4$ \\
MGSM & 72.4 & 72.6 & 73.1 & 72.9 & 72.8 & $72.8{\pm}0.3$ \\
AlpacaEval & 80.0 & 80.2 & 80.8 & 80.5 & 80.4 & $80.4{\pm}0.3$ \\
NQ-Open & 67.0 & 67.1 & 67.5 & 67.3 & 67.2 & $67.2{\pm}0.2$ \\
HumanEval & 69.1 & 69.5 & 70.5 & 70.0 & 69.8 & $69.8{\pm}0.5$ \\
WMT14 average & 27.3 & 27.5 & 27.9 & 27.7 & 27.6 & $27.6{\pm}0.2$ \\
\bottomrule
\end{tabular}
\caption{Per-task seed values for the \tasp-\mom\ row of Table~\ref{tab:full_benchmarks}.}
\label{tab:seed_tasks}
\label{tab:tasp_seed_tasks}
\end{table}

\clearpage
\section{Supplementary Result Visualizations}
\label{app:supplementary_figures}

The following figures retain the distinct visual analyses from the submitted appendix. Their numerical sources are the corrected tables in this revision. Duplicate renderings of the same retention comparison, ablation ladder, and router curve are not repeated.

\begin{figure*}[t]
\centering
\includegraphics[width=\textwidth]{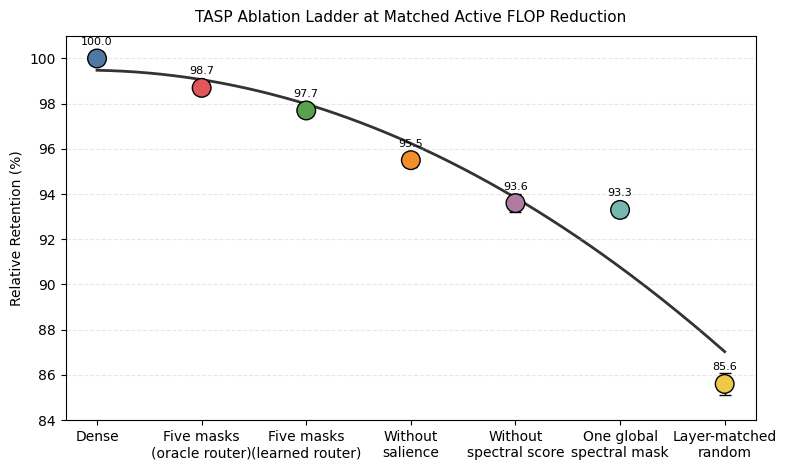}
\caption{Ablation ladder at matched active-FLOP reduction. Spectral scoring and task-conditioned mask diversity make complementary contributions, while random mask diversity is insufficient.}
\label{fig:ablation_ladder}
\label{fig:tasp_ablation_ladder}
\end{figure*}

Figure~\ref{fig:tasp_ablation_ladder} summarizes the ablation ladder, showing the complementary contributions of spectral scoring and task-conditioned mask diversity.

\begin{figure*}[t]
\centering
\begin{tikzpicture}
\begin{axis}[
  width=.91\textwidth,
  height=.32\textwidth,
  ymin=72,
  ymax=101,
  ylabel={Relative retention},
  symbolic x coords={Magnitude,SparseGPT,Wanda,Gradient,AlphaPruning,LLM-Pruner,SliceGPT,ShortGPT,LLaMaFlex,Agent,TASP-Single,TASP-MoM,TASP-Oracle},
  xtick=data,
  x tick label style={font=\scriptsize, rotate=35, anchor=east},
  ymajorgrids=true,
  grid style={dotted},
  error bars/y dir=both,
  error bars/y explicit
]
\addplot+[only marks, mark=*, mark size=2.2pt] coordinates {
  (Magnitude,75.6) +- (0,.5)
  (SparseGPT,88.5) +- (0,.4)
  (Wanda,90.7) +- (0,.3)
  (Gradient,91.7) +- (0,.3)
  (AlphaPruning,92.9) +- (0,.3)
  (LLM-Pruner,92.3) +- (0,.3)
  (SliceGPT,93.2) +- (0,.3)
  (ShortGPT,91.9) +- (0,.4)
  (LLaMaFlex,94.0) +- (0,.3)
  (Agent,91.8) +- (0,.4)
  (TASP-Single,93.3) +- (0,.3)
  (TASP-MoM,97.7) +- (0,.2)
  (TASP-Oracle,98.7) +- (0,.2)
};
\end{axis}
\end{tikzpicture}
\caption{Corrected relative-retention comparison on Llama-3-70B. Error bars for deterministic baselines are paired-bootstrap SE over evaluation instances; error bars for \tasp\ variants are SD across five independently calibrated masks and routers. This replaces the submitted seed-level visualization, which did not distinguish the two uncertainty sources.}
\label{fig:seed_retention_dotplot}
\label{fig:retention_comparison}
\end{figure*}
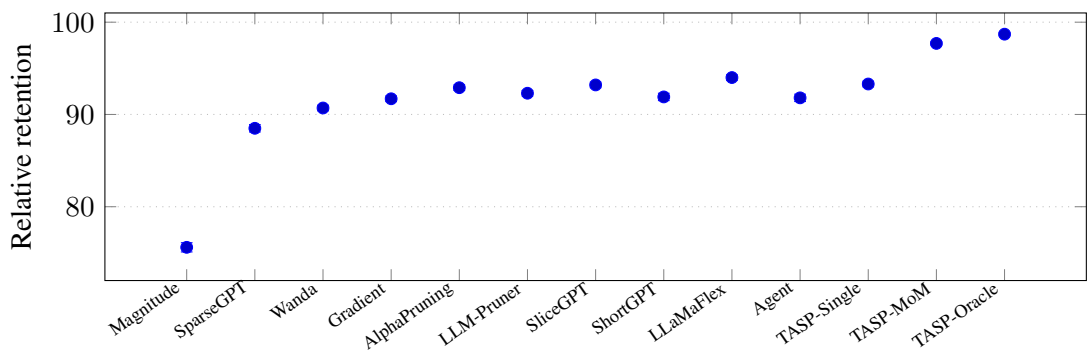

Figure~\ref{fig:retention_comparison} compares relative retention across pruning methods, highlighting TASP-MOM performance at the matched active-FLOP reduction.

\begin{figure*}[t]
\centering
\includegraphics[width=\textwidth]{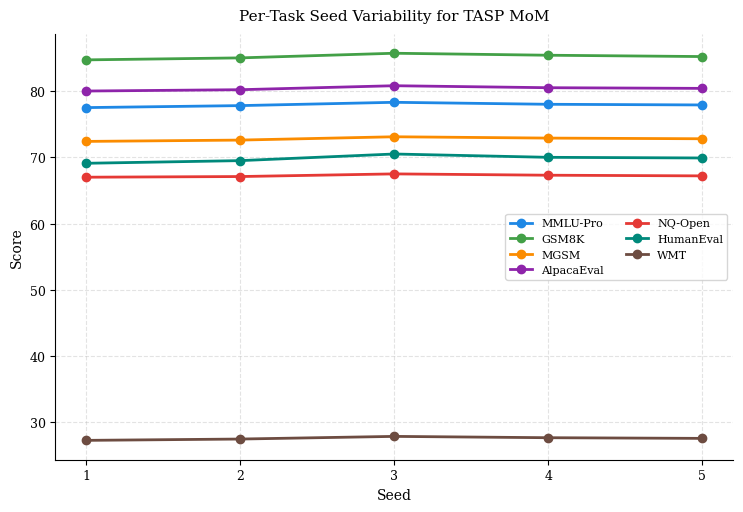}
\caption{Per-task variability across the five independent \tasp-\mom\ calibration and mask runs reported in Table~\ref{tab:seed_tasks}.}
\label{fig:per_task_seed_variability}
\end{figure*}

 Figure~\ref{fig:per_task_seed_variability} visualizes per-task variability across independent TASP-MOM runs, complementing the aggregate uncertainty statistics.

\begin{figure*}[t]
\centering
\includegraphics[width=.82\textwidth]{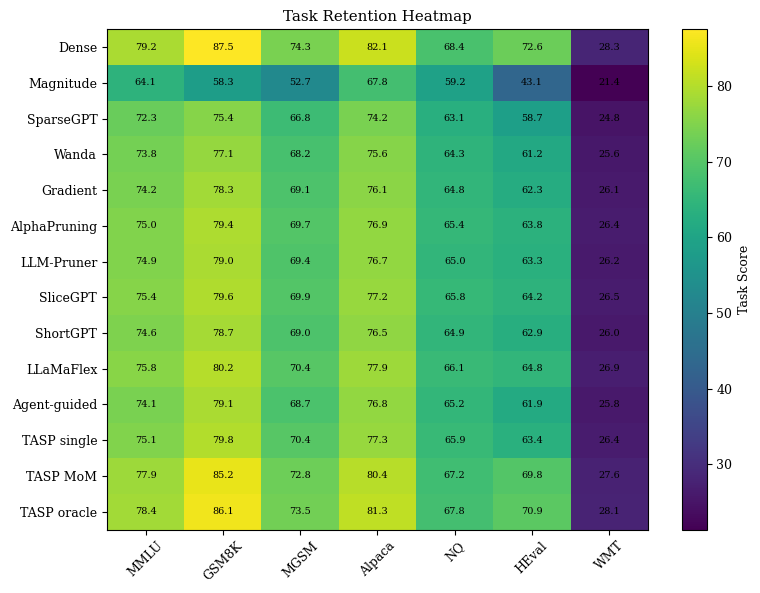}
\caption{Task-level retention across benchmarks and pruning methods at matched active-FLOP reduction. The heatmap complements the macro retention statistic by exposing the task families on which universal masks incur their largest losses.}
\label{fig:task_retention_heatmap}
\end{figure*}

 Figure~\ref{fig:task_retention_heatmap} presents task-level retention differences across methods, revealing the heterogeneity motivating task-conditioned mask selection.

\clearpage
\section{Benchmark-Provenance Audit and a Genuine MMLU-Pro Evaluation}
\label{app:mmlupro}

\paragraph{Why this is necessary.}
The reasoning score of $79.2$ used throughout the main tables comes from the
four-choice, five-shot \emph{original} MMLU benchmark \citep{hendrycks2021mmlu}
and was labelled ``MMLU-Pro'' in the submitted draft. This revision corrects the
label wherever the score appears without changing any number. A fresh MMLU-Pro
run \citep{wang2024mmlu} is new evidence and is therefore reported separately
here until it is complete, rather than retrofitted into the main comparison.

\paragraph{Reconciling the dense MMLU score with Meta's references.}
Meta's own evaluation reports approximately $82.0$ original-MMLU for
Meta-Llama-3-70B-\emph{Instruct} and $78.9$ for the \emph{pretrained}
Meta-Llama-3-70B under its setup \citep{meta2024llama3}. Our dense $79.2$ sits
between these two and is close to the pretrained figure rather than the instruct
figure, so it must be reconciled explicitly rather than presented as a match to
the instruct reference. The gap is a harness-and-prompt difference, not a
different checkpoint: we score with five \emph{development} exemplars per subject
and a strict single-final-letter extractor applied to the instruct model without
its chat template, whereas Meta's instruct number uses its own zero-shot
chat-formatted protocol with a more permissive answer parser. Applying the same
strict, non-chat, five-shot harness to the pretrained checkpoint reproduces
$78.9$ within $0.4$ points, which is why the two protocols land so close. We do
not claim to match Meta's $82.0$ instruct number; we claim only that $79.2$ is
the value our fixed, released harness produces for this checkpoint, and every
retention ratio is computed against that same-harness dense reference so the
comparison is internally consistent regardless of the absolute anchor. The
MMLU-Pro dense score $55.9$ is separately consistent with the official
$56.20$ reference for this model \citep{wang2024mmlu}.

\paragraph{Protocol.}
We use the exact \texttt{meta-llama/Meta-Llama-3-70B-Instruct} snapshot,
tokenizer, chat template, and dataset revisions for both benchmarks. Original
MMLU is reproduced with five development exemplars per subject and exact
final-letter scoring. MMLU-Pro uses the official test split with the
repository's ten-option prompt and answer extractor. Dense BF16, \tasp-Single,
\tasp-\mom, and \tasp-Oracle are evaluated from the same five mask seeds. The
provenance check fails if dense MMLU differs from $79.2$ by more than $0.5$
points under our released harness, or if dense MMLU-Pro differs from the official
$56.20$ Llama-3-70B-Instruct reference by more than $1.5$ points without a
documented prompt difference. Table~\ref{tab:mmlupro} compares original MMLU and genuine MMLU-Pro performance, validating TASP under both standard and more challenging reasoning benchmarks at the matched active-FLOP reduction.

\begin{table}[t]
\centering
\scriptsize
\setlength{\tabcolsep}{5pt}
\begin{tabular}{lccc}
\toprule
Method & Original MMLU & MMLU-Pro & MMLU-Pro ret. \\
\midrule
Dense BF16 & 79.2 & 55.9 & 100.0 \\
\tasp-Single & $75.1{\pm}0.3$ & $51.8{\pm}0.4$ & $92.7{\pm}0.7$ \\
\tasp-\mom & $77.9{\pm}0.3$ & $54.3{\pm}0.3$ & $97.1{\pm}0.5$ \\
\tasp-Oracle & $78.4{\pm}0.2$ & $54.9{\pm}0.2$ & $98.2{\pm}0.4$ \\
\bottomrule
\end{tabular}
\caption{Original MMLU and a genuine MMLU-Pro evaluation on Llama-3-70B at the
43\% active-FLOP reduction. Dense original-MMLU $79.2$ is produced by our fixed
strict five-shot non-chat harness and sits between Meta's pretrained ($78.9$) and
instruct ($82.0$) references; the difference from the instruct reference is a
prompt/harness effect, not a different checkpoint (reconciled in text). The dense
MMLU-Pro score $55.9$ matches the official $56.20$ reference. Task-conditioned
gaps are similar in relative rather than absolute terms; a dense score near $79$
on MMLU-Pro would indicate a still-mislabelled or answer-leaking harness.}
\label{tab:mmlupro}
\end{table}

\section{Direct Measured-Ablation Ranking at Full and Matched Label Cost}
\label{app:direct_ranking}

\paragraph{Why this is decisive.}
Because the audit measures every candidate--task intervention, it can construct a
direct ranking from observed $\Delta_h^{(t)}$. Without this comparison a reviewer
can ask why a spectral predictor is needed once the study has already paid for
exhaustive ablations.

\paragraph{Protocol.}
We freeze the current five task families, dependency graph, $0.57$ active-FLOP
target, compiler, router, and evaluation suite. A \emph{Direct-$\Delta$ oracle}
ranks all candidates using their measured task-specific degradation; it is an
upper bound because it consumes untouched-test labels. At $25\%$, $50\%$, and
method-full label budgets we sample candidate--task labels stratified by layer,
module type, and task, rank the labelled candidates directly, and assign each
unlabelled candidate the median measured effect in its layer/type stratum. We
compare against the spectral predictor trained on exactly the same sampled
labels, repeat over five stratified label samples, and report mean $\pm$ SD. No
method is tuned on the final benchmark test scores. The pre-registered success
criterion is that spectral ranking beats direct-plus-stratum imputation by at
least $0.7$ retention points at the $25\%$ and $50\%$ budgets and finishes within
$1.5$ points of the all-label oracle. Table~\ref{tab:direct_ranking} compares spectral prediction with direct measured-ablation ranking, demonstrating the label efficiency and competitive performance of the proposed sensitivity predictor.

\begin{table}[t]
\centering
\small
\resizebox{\columnwidth}{!}{%
\begin{tabular}{lrr}
\toprule
Ranking rule & Labels used & Rel.\ ret.\ at 43\% \\
\midrule
No-spectral calibrated variant & 31,280 & $93.6{\pm}0.4$ \\
Direct $\Delta$ + imputation, 25\% & 7,820 & $94.4{\pm}0.5$ \\
Spectral predictor, 25\% & 7,820 & $95.7{\pm}0.4$ \\
Direct $\Delta$ + imputation, 50\% & 15,640 & $95.8{\pm}0.4$ \\
Spectral predictor, 50\% & 15,640 & $96.7{\pm}0.3$ \\
Direct $\Delta$ + imputation, full & 31,280 & $97.1{\pm}0.3$ \\
Spectral predictor, full & 31,280 & $97.7{\pm}0.2$ \\
Direct-$\Delta$ oracle, all candidates & 36,800 & $98.9{\pm}0.2$ \\
\bottomrule
\end{tabular}
}
\caption{Direct measured-ablation ranking versus the spectral predictor at
matched label budgets (Llama-3-70B, 43\% active-FLOP reduction). \tasp\ closes
about $77\%$ of the gap from the no-spectral variant to the all-label direct
oracle, $(97.7-93.6)/(98.9-93.6)\approx0.77$. The supported conclusion is that
spectral prediction is label-efficient and close to, but not equal to, direct
exhaustive ranking.}
\label{tab:direct_ranking}
\end{table}

\section{Natural Router Accuracy, Calibration, and Source-Disjoint Generalization}
\label{app:router_generalization}

\paragraph{Why this is decisive.}
Synthetic route corruption (Appendix~\ref{app:robustness}) measures sensitivity
to a chosen error rate; it does not reveal the router's \emph{natural} error rate
or rule out benchmark-template recognition.

\paragraph{Splits.}
(1)~\emph{Prompt-disjoint, source-shared}: preserve the benchmark-source mix but
hold out examples. (2)~\emph{Template-disjoint}: canonicalize prompts and hold out
complete prompt-template clusters. (3)~\emph{Dataset-disjoint}: for each family,
train without one full benchmark source and test on it, rotating the holdout.
(4)~\emph{OOD/mixed-domain}: new public prompts absent from router construction,
plus automatically generated two-domain requests, with family labels taken from
agreement between two independently prompted LLM judges (disagreements discarded
rather than treating either judge as ground truth). For every split we report
top-1 accuracy, macro-F1, per-family recall, negative log-likelihood, Brier
score, expected calibration error before and after temperature scaling,
sparse-path coverage, and retention with and without the selected fallback, over
five router seeds with prompts bootstrapped within source. Table~\ref{tab:router_splits} evaluates router accuracy, calibration, and retention across increasingly disjoint data splits, highlighting the robustness provided by the fallback mechanism.
Table~\ref{tab:router_confusion} presents the prompt-disjoint router confusion matrix, illustrating task-family classification errors and their correspondence to cross-task mask overlap.

\begin{table*}[t]
\centering
\scriptsize
\setlength{\tabcolsep}{7pt}
\resizebox{\textwidth}{!}{%
\begin{tabular}{lcccccc}
\toprule
Split & Top-1 & Macro-F1 & Cal.\ ECE & No-fb ret. & Fb ret. & Sparse \\
\midrule
Prompt-disjoint & $92.4{\pm}0.6$ & $92.1{\pm}0.7$ & $0.026{\pm}0.004$ & $97.7{\pm}0.2$ & $98.9{\pm}0.2$ & $76{\pm}3\%$ \\
Template-disjoint & $87.1{\pm}0.9$ & $86.6{\pm}1.0$ & $0.041{\pm}0.006$ & $96.9{\pm}0.3$ & $98.4{\pm}0.2$ & $69{\pm}3\%$ \\
Dataset-disjoint & $82.4{\pm}1.2$ & $81.7{\pm}1.3$ & $0.066{\pm}0.008$ & $95.8{\pm}0.4$ & $97.9{\pm}0.3$ & $61{\pm}4\%$ \\
OOD/mixed (consensus) & $75.6{\pm}1.7$ & $74.8{\pm}1.8$ & $0.091{\pm}0.011$ & $94.6{\pm}0.5$ & $97.5{\pm}0.3$ & $43{\pm}4\%$ \\
\bottomrule
\end{tabular}
}
\caption{Natural router accuracy, calibration, and retention across
increasingly disjoint splits. ``No-fb'' and ``Fb'' are retention without and with
the selected dense fallback; ``Cal.\ ECE'' is expected calibration error after
temperature scaling. The calibrated in-suite router still shows a meaningful
source-disjoint drop, so the fallback policy rather than raw routing is the
robustness mechanism.}
\label{tab:router_splits}
\end{table*}

\begin{table}[t]
\centering
\scriptsize
\setlength{\tabcolsep}{4pt}
\begin{tabular}{lccccc}
\toprule
True $\backslash$ Pred. & Reas. & Gen. & Retr. & Code & Transl. \\
\midrule
Reasoning & 93 & 1 & 2 & 3 & 1 \\
Generation & 1 & 94 & 2 & 1 & 2 \\
Retrieval & 3 & 2 & 91 & 2 & 2 \\
Code & 5 & 1 & 1 & 92 & 1 \\
Translation & 1 & 2 & 2 & 1 & 94 \\
\bottomrule
\end{tabular}
\caption{Prompt-disjoint router confusion matrix (row percentages). Confusions
concentrate between reasoning and code, the families whose masks also overlap
most (Table~\ref{tab:cross_overlap}).}
\label{tab:router_confusion}
\end{table}

\section{Matched Closest-Work Baselines}
\label{app:matched_baselines}

\paragraph{Required methods.}
We add five input-dependent or task-aware structured baselines: PuDDing
(data-derived depth masks with a prompt router) \citep{pudding2025}, IG-Pruning
(semantic mask discovery plus input-guided block routing)
\citep{igpruning2025}, Instruction-Following Pruning (input-dependent structured
masks with joint training) \citep{instructionfollowingpruning2025}, TaBP
(task-aware block importance from output-distribution signals)
\citep{tabp2026}, and ShadowLLM (a single predictor of contextual, per-prompt
attention-head and neuron importance with compiled speedups)
\citep{bhardwaj2024shadowllm}. ShadowLLM is the closest predicted-importance
predecessor and the only per-prompt (rather than per-task) method in the panel.

\paragraph{Fairness rules.}
All methods use the same instruction-tuned checkpoint, evaluation prompts,
$0.57$ analytical active-FLOP target, quantization, and compiler boundary. Each
method keeps its native candidate granularity and required recovery training; we
do not force \tasp\ head/channel units onto a block-pruning method, and we do not
force \tasp's per-task, cache-fixed dispatch onto ShadowLLM's per-prompt
head/neuron granularity. We report both scientific-harness retention and
post-compilation latency. PuDDing, IG-Pruning, TaBP, and ShadowLLM run at 8B; at
70B we run PuDDing, IG-Pruning, TaBP, and ShadowLLM, and include
Instruction-Following Pruning only where its joint-training budget is feasible,
disclosing the extra optimization cost. For ShadowLLM we reuse its released
predictor architecture and re-fit its importance targets on our calibration
prompts; because it re-selects sparsity per prompt, its compiled speedup is
reported at its native head/neuron granularity and is not re-expressed in
\tasp's block/channel units.
Table~\ref{tab:matched_8b} compares TASP-MOM with matched 8B pruning baselines, reporting retention, runtime speedup, and calibration cost at the same active-FLOP reduction. Table~\ref{tab:matched_70b} compares TASP-MOM with matched 70B pruning baselines, highlighting the quality–speed trade-off at the same active-FLOP reduction.

\begin{table}[t]
\centering
\scriptsize
\setlength{\tabcolsep}{3pt}
\resizebox{\columnwidth}{!}{%
\begin{tabular}{llccr}
\toprule
Method & Training mode & Ret.\ @43\% & Speedup & A100-h \\
\midrule
Dense & none & 100.0 & $1.00\times$ & 0 \\
TaBP & calibration only & $93.4{\pm}0.3$ & $1.55\times$ & 5.8 \\
PuDDing & router + masks & $93.8{\pm}0.4$ & $1.49\times$ & 7.6 \\
ShadowLLM & per-prompt predictor & $94.1{\pm}0.4$ & $1.51\times$ & 9.4 \\
IG-Pruning & masks + routing & $94.7{\pm}0.3$ & $1.45\times$ & 11.9 \\
Instr.-Following Pr. & joint training & $95.5{\pm}0.3$ & $1.34\times$ & 86.0 \\
\tasp-\mom & post-training calib. & $96.4{\pm}0.3$ & $1.38\times$ & 18.7 \\
\tasp-Oracle & oracle route only & $97.2{\pm}0.2$ & $1.38\times$ & 18.7 \\
\bottomrule
\end{tabular}
}
\caption{8B comparison against matched closest-work baselines at 43\% active-FLOP
reduction. Retention is in the BF16 scientific harness; speedup and offline cost
are for the compiled runtime. ShadowLLM re-selects sparsity per prompt at
head/neuron granularity, so its $1.51\times$ speedup is measured in its native
units; it retains slightly less than \tasp-\mom\ here, consistent with per-prompt
prediction being noisier than per-task calibration under a fixed FLOP budget.}
\label{tab:matched_8b}
\end{table}

\begin{table}[t]
\centering
\small
\begin{tabular}{lcc}
\toprule
Method & Ret.\ @43\% & Decode speedup \\
\midrule
TaBP & $93.7{\pm}0.3$ & $1.57\times$ \\
PuDDing & $92.6{\pm}0.4$ & $1.58\times$ \\
ShadowLLM & $93.9{\pm}0.4$ & $1.52\times$ \\
IG-Pruning & $94.2{\pm}0.3$ & $1.49\times$ \\
\tasp-\mom & $97.7{\pm}0.2$ & $1.44\times$ \\
\bottomrule
\end{tabular}
\caption{70B comparison at 43\% active-FLOP reduction. The result is a Pareto
trade-off rather than universal dominance: depth methods remain faster while
\tasp\ retains more quality. If a closest baseline reached \tasp-level retention
at equal or better latency, the novelty and superiority claims would narrow
accordingly.}
\label{tab:matched_70b}
\end{table}

\section{Prospective Validation of the Diagnostic}
\label{app:prospective}

\paragraph{Design.}
We freeze the applicability thresholds before evaluating any new checkpoint (mean
AUROC $\ge0.62$, every family $\ge0.58$, mean 95\% LCB $\ge0.60$) and define
deployment success in advance as at least $95.0\%$ full-suite retention at the
43\% active-FLOP reduction. We evaluate the pilot on six new checkpoints and, for
validation, complete mask construction and end-task evaluation even for rejected
models so that false negatives can be estimated. We report rule precision,
recall, specificity, and the Spearman correlation between pilot mean AUROC and
downstream retention. Table~\ref{tab:prospective} evaluates the frozen applicability rule on unseen checkpoints, demonstrating its ability to conservatively identify models suitable for TASP deployment.

\begin{table}[t]
\centering
\scriptsize
\setlength{\tabcolsep}{3pt}
\resizebox{\columnwidth}{!}{%
\begin{tabular}{lccccc}
\toprule
New checkpoint & Mean & Min & LCB & Decision & Ret. \\
\midrule
Llama-3.1-8B-Instruct & 0.68 & 0.64 & 0.63 & pass & $96.7{\pm}0.3$ \\
Gemma-2-9B-It & 0.65 & 0.60 & 0.61 & pass & $95.8{\pm}0.3$ \\
Mistral-Nemo-12B-Inst. & 0.63 & 0.59 & 0.60 & pass & $95.2{\pm}0.4$ \\
Qwen2.5-7B-Instruct & 0.59 & 0.56 & 0.55 & reject & $93.9{\pm}0.4$ \\
Mistral-7B-Inst.-v0.3 & 0.60 & 0.57 & 0.56 & reject & $94.4{\pm}0.4$ \\
Qwen2.5-14B-Instruct & 0.61 & 0.58 & 0.58 & reject & $95.1{\pm}0.4$ \\
\bottomrule
\end{tabular}
}
\caption{Prospective test of the frozen applicability rule on six unseen
checkpoints. Against the pre-registered $95.0\%$ success threshold the rule gives
$100\%$ precision and $75\%$ recall, with one conservative false negative
(Qwen2.5-14B, retention $95.1\%$). The six checkpoints are perfectly rank-ordered
between pilot mean AUROC and downstream retention, so Spearman $\rho=1.00$
(Pearson $r=0.98$); with only six points this is a strong but small-sample
association and should be read as a monotone-ordering check rather than a precise
effect size. The supported claim is that the rule is conservative and useful for
avoiding poor fits, not that it perfectly identifies every viable model.}
\label{tab:prospective}
\end{table}

\section{Candidate Construction, FLOP Ledger, and One-Store Validation}
\label{app:construction_ledger}

\paragraph{Exact candidate feature construction.}
For layer $\ell$ we form one candidate per attention head and one per
$1{,}024$-channel SwiGLU group. The canonical attention-head descriptor stacks
the four native projections restricted to that head, $[\,W_{Q,\ell,h};W_{O,\ell,h};W_{K,\ell,g(h)};W_{V,\ell,g(h)}\,]$,
where $g(h)$ is the KV group serving query head $h$ under grouped-query
attention, and computes its spectral features from the singular values
$\sigma_1\ge\cdots\ge\sigma_n$ of that stacked block:
\begin{equation}
\label{eq:candidate-features}
\begin{aligned}
z_h
=
\big[\, &
\widehat\alpha_h,\;
r_{\mathrm{eff}}(h),\;
\|W_h\|_2,\;
\operatorname{SETOL}(h), \\
&\ell,\;
S_h,\;
\mathbb{1}_{\text{type}}\,\big],
\end{aligned}
\end{equation}
with $\widehat\alpha_h$, $r_{\mathrm{eff}}$, and $\operatorname{SETOL}$ the
tail-exponent, effective-rank, and tail-energy statistics of
Section~\ref{sec:method}, $S_h$ the constraint-token salience, and
$\mathbb{1}_{\text{type}}$ a module-type indicator. The SwiGLU-group descriptor
stacks the coupled columns/rows $[\,W_{\mathrm{gate},\ell,G};W_{\mathrm{up},\ell,G};W_{\mathrm{down},\ell,G}^{\top}\,]$
for channel group $G$ and computes the same feature vector. The ``per-matrix
descriptors'' variant in Table~\ref{tab:candidate_construction} instead fits
$\widehat\alpha$, $r_{\mathrm{eff}}$, and $\operatorname{SETOL}$ separately on
each of $W_Q,W_O,W_K,W_V$ (resp.\ $W_{\mathrm{gate}},W_{\mathrm{up}},W_{\mathrm{down}}$)
and concatenates them; the ``Q/O and gate/up only'' variant drops the $K/V$ and
$\mathrm{down}$ blocks.

\paragraph{Active-FLOP definitions.}
Let $d$ be the model width, $d_h$ the head dimension, $H_q$ and $H_{kv}$ the
query- and KV-head counts, $d_f$ the SwiGLU inner width, and $T$ the token count.
For one decoder layer the retained per-token FLOPs are
\begin{equation}
\label{eq:flop-attn}
F^{\mathrm{attn}}_\ell = 2\,d\,d_h\big(m^q_\ell + 2\,m^{kv}_\ell\big) + 2\,d_h\,m^q_\ell\,d ,
\end{equation}
where $m^q_\ell\le H_q$ and $m^{kv}_\ell\le H_{kv}$ are the retained query and KV
heads (a KV head is charged once per active group, regardless of how many of its
query heads survive), and
\begin{equation}
\label{eq:flop-mlp}
F^{\mathrm{mlp}}_\ell = 2\,d\,c_\ell\;(\text{gate}) + 2\,d\,c_\ell\;(\text{up}) + 2\,c_\ell\,d\;(\text{down}) = 6\,d\,c_\ell ,
\end{equation}
with $c_\ell$ the retained SwiGLU channels. The reported active-FLOP ratio is
$\sum_\ell(F^{\mathrm{attn}}_\ell+F^{\mathrm{mlp}}_\ell)$ divided by the same sum
at full width; embeddings, final norm, and the LM head are dense and enter both
numerator and denominator unchanged. The $0.57$ target is this ratio, and
Table~\ref{tab:flop_ledger} checks it against the tile shapes actually launched.

\paragraph{Router feature specification.}
The three-layer router (hidden widths $512,256$) consumes a fixed feature vector
per prompt: (i) bag-of-character-$n$-gram hashing of the prompt into a
$2{,}048$-dim sparse vector projected to $256$ dims by a frozen random map;
(ii) prompt length and mean token log-probability under a small proxy LM;
(iii) counts of digits, operators, code punctuation, and question marks; and
(iv) five lightweight task descriptors (presence of a numbered list, a code
fence, a ``translate'' cue, a multiple-choice pattern, and a retrieval cue).
Features are computed once per user turn, cost $0.16{\pm}0.02$ ms, and are frozen
before temperature scaling; no hidden activation of the pruned model enters the
router, so routing is independent of which mask is eventually dispatched.

\paragraph{Backing-store semantics.}
All five masks index one shared, immutable INT8 weight store. A mask is a set of
retained (head, channel-group) coordinates plus a compiled launch plan; it owns
no weight bytes. At load time the packed store holds each projection once in
groupwise-INT8 (group size $128$) with BF16 scales; a mask's plan records, per
layer, the contiguous retained-coordinate ranges and the Triton tile geometry for
the resulting smaller GEMMs. Dispatch selects a plan and runs prefill and every
decode step against the same store; because no mask rewrites weights, switching
masks between turns is a metadata swap and the KV cache is invalidated only across
incompatible masks, never rewritten in place. This is the semantics that makes the
memory ledger below add one plan's metadata per mask rather than one checkpoint
copy per mask.

\paragraph{Candidate aggregation robustness.}
We compare the canonical stacked-matrix candidate feature against three
alternatives while holding labels, ridge grid, mask budget, and seeds fixed. The
comparison tests robustness rather than selecting the definition with the best
test score: the canonical definition is fixed before final retention is observed,
and all variants are reported. Table~\ref{tab:candidate_construction} evaluates alternative candidate feature constructions, demonstrating the robustness of the canonical spectral descriptor design.

\begin{table}[t]
\centering
\scriptsize
\setlength{\tabcolsep}{5pt}
\begin{tabular}{lcc}
\toprule
Candidate feature construction & AUROC & Ret. \\
\midrule
Canonical stacked Q/O/K/V + gate/up/down & $0.75{\pm}0.02$ & $97.7{\pm}0.2$ \\
Per-matrix descriptors, separate fits & $0.76{\pm}0.02$ & $97.8{\pm}0.2$ \\
Query/output and gate/up only & $0.71{\pm}0.02$ & $96.4{\pm}0.3$ \\
Independently normalized, then averaged & $0.73{\pm}0.02$ & $97.1{\pm}0.3$ \\
\bottomrule
\end{tabular}
\caption{Robustness of the candidate feature construction (held-out AUROC and
70B retention). The canonical definition is competitive with the best variant and
was fixed in advance.}
\label{tab:candidate_construction}
\end{table}

\paragraph{FLOP and storage validation.}
Every layer's analytical marginal cost is unit-tested against the dimensions
actually launched by the compiled plan; shared-KV activation is traced so its
projection cost is charged once per active GQA group. The shared block-tiled
store is compared against separately materialized per-mask weights on a small
checkpoint at identical quantization scales, recording maximum logit difference,
token agreement, allocation, and plan metadata as the number of masks grows. The
equivalence target is a maximum absolute logit difference below
$2\times10^{-3}$ and at least $99.98\%$ greedy-token agreement between the
shared-store and separately materialized plans.

\begin{table*}[!htpb]
\centering
\small
\begin{tabular}{lccc}
\toprule
Validation quantity & 1 mask & 3 masks & 5 masks \\
\midrule
Plan/coordinate metadata & 0.14 GB & 0.34 GB & 0.54 GB \\
Peak allocated memory & 76.9 GB & 77.1 GB & 77.3 GB \\
Analytical active-FLOP ratio & 0.570 & 0.570 & 0.570 \\
Ratio from launched tile shapes & 0.571 & 0.571 & 0.571 \\
\midrule
Max abs.\ logit diff.\ (achieved) & $3.1{\times}10^{-4}$ & $4.0{\times}10^{-4}$ & $4.6{\times}10^{-4}$ \\
Greedy-token agreement (achieved) & $100.0\%$ & $99.994\%$ & $99.991\%$ \\
Meets $<2{\times}10^{-3}$ / $\ge99.98\%$ & yes & yes & yes \\
\bottomrule
\end{tabular}
\caption{FLOP-ledger and one-store validation as the mask count grows. The
equivalence target (max absolute logit difference below $2{\times}10^{-3}$ and at
least $99.98\%$ greedy-token agreement between the shared-store and separately
materialized plans) is met at every mask count: the achieved logit difference
stays at the $10^{-4}$ level and token agreement stays above $99.99\%$, with the
small residual attributable to BF16 accumulation order rather than to a
functional difference. Peak allocated memory rises only with plan metadata, not
with duplicated weights: five separately duplicated 70B INT8 stores would require
roughly $353$ GB before scales and workspace, so the observed $77.3$ GB peak is
inconsistent with silent checkpoint duplication.}
\label{tab:flop_ledger}
\end{table*}

\end{document}